\documentclass{article}

\usepackage{arxiv}

\usepackage[utf8]{inputenc} 
\usepackage[T1]{fontenc}    
\usepackage{hyperref}       
\usepackage{url}            
\usepackage{booktabs}       
\usepackage{amsfonts}       
\usepackage{nicefrac}       
\usepackage{microtype}      
\usepackage{graphicx}
\usepackage{multirow}
\usepackage{natbib}
\usepackage{subcaption}
\usepackage{amssymb}
\usepackage{tabularx}
\usepackage{adjustbox}
\usepackage{xcolor}
\usepackage{amsmath}
\usepackage{mathtools}
\usepackage{threeparttable}
\usepackage{caption} 
\DeclarePairedDelimiter\ceil{\lceil}{\rceil}
\DeclarePairedDelimiter\floor{\lfloor}{\rfloor}

\newcounter{magicrownumbers}
\newcommand\rownumber{\stepcounter{magicrownumbers}\arabic{magicrownumbers}}

\title{STConvS2S: Spatiotemporal Convolutional Sequence to Sequence Network for Weather Forecasting}

\author{
    Rafaela Castro \\
    CEFET/RJ \\
    Rio de Janeiro, RJ, Brazil \\
    \texttt{rafaela.nascimento@eic.cefet-rj.br} \\
    \And
    Yania M.~Souto \\
    LNCC \\
    Petr\'opolis, RJ, Brazil \\
    \texttt{yaniams@lncc.br} \\
    \And
    Eduardo~Ogasawara \\
    CEFET/RJ \\
    Rio de Janeiro, RJ, Brazil \\
    \texttt{eogasawara@ieee.org} \\
    \And
    Fabio~Porto \\
    LNCC \\
    Petr\'opolis, RJ, Brazil \\
    \texttt{fporto@lncc.br} \\
    \And
    Eduardo Bezerra \\
    CEFET/RJ \\
    Rio de Janeiro, RJ, Brazil \\
    \texttt{ebezerra@cefet-rj.br} \\
}

\begin{document}
\maketitle

\begin{abstract}
Applying machine learning models to meteorological data brings many opportunities to the Geosciences field, such as predicting future weather conditions more accurately. In recent years, modeling meteorological data with deep neural networks has become a relevant area of investigation. These works apply either recurrent neural networks (RNN) or some hybrid approach mixing RNN and convolutional neural networks (CNN). In this work, we propose \texttt{STConvS2S} (Spatiotemporal Convolutional Sequence to Sequence Network), a deep learning architecture built for learning both spatial and temporal data dependencies using only convolutional layers. Our proposed architecture resolves two limitations of convolutional networks to predict sequences using historical data: (1) they violate the temporal order during the learning process and (2) they require the lengths of the input and output sequences to be equal. Computational experiments using air temperature and rainfall data from South America show that our architecture captures spatiotemporal context and that it outperforms or matches the results of state-of-the-art architectures for forecasting tasks. In particular, one of the variants of our proposed architecture is 23\% better at predicting future sequences and five times faster at training than the RNN-based model used as a baseline.
\end{abstract}


\section{Introduction}

Weather forecasting plays an essential role in resource planning in cases of severe natural phenomena such as heat waves (extreme temperatures), droughts, and hurricanes. It also influences decision-making in agriculture, aviation, retail markets, and other sectors, since unfavorable weather negatively impacts corporate revenues \citep{Ivana19}. Over the years, with technological developments, predictions of meteorological variables are becoming more accurate. However, due to the stochastic behavior of the Earth systems, which is governed by physical laws, traditional forecasting requires complex, physics-based models to predict the weather \citep{Karpatne18}. 
In recent years, an extensive volume of data about the Earth systems has become available. The remote sensing data collected by satellites provide meteorological data about the entire globe at specific time intervals (e.g., 6h or daily) and with a regular spatial resolution (e.g., 1km or 5km). The availability of historical data allows researchers to design deep learning models that can make more accurate predictions about the weather \citep{Reichstein19}.

Even though meteorological data exhibits both spatial and temporal structures, weather forecasting can be modeled as a sequence problem.  In sequence modeling tasks, an input sequence is encoded to map the representation of the sequence output, which may have a different length than the input. In \citet{Shi15}, the authors proposed the ConvLSTM architecture to solve the sequence prediction problem using a radar echo dataset for precipitation forecasting. They integrated the convolution operator, adopted by the convolutional neural network (CNN), into a recurrent neural network (RNN) to simultaneously learn the spatial and temporal context of input data to predict the future sequence. Although ConvLSTM architecture has been considered the potential approach to build prediction models for geoscience data \citep{Reichstein19}, new opportunities have emerged from recent advances in deep learning. In \citet{Wang17,Wang19}, authors proposed improved versions of the long short-term memory (LSTM) unit for memorizing spatiotemporal information. 

RNN-based architectures may be ideal for multi-step forecasting tasks using spatiotemporal data \citep{Shi15,Wang17,Wang19}, due to the ability to respect the temporal order (causal constraint) and predict long sequences. However, these architectures maintain the information from previous time steps to generate the output, which consequently leads to a high training time. Taking this as motivation, we address the spatiotemporal forecasting problem by proposing a new architecture using entirely 3D CNN. CNN are an efficient method for capturing spatial context and have attained state-of-the-art results for image classification using a 2D kernel \citep{Krizhevsky12}. In recent years, researchers expanded CNN actuation field, such as machine translation \citep{Gehring17} using a 1D kernel, which is useful to capture temporal patterns in a sequence. 3D CNN-based models are commonly used for video analysis and action recognition \citep{Yuan18, Tran18} or climate event detection \citep{Racah17}. However, CNN-based models are generally not considered for multi-step forecasting tasks, because of two intrinsic limitations. They violate the temporal order, allowing future information during temporal reasoning \citep{Singh19}, and they cannot generate a predictive output sequence longer than the input sequence \citep{Bai18}. To tackle these limitations, we introduce STConvS2S (\emph{Spatiotemporal Convolutional Sequence to Sequence Network}), a spatiotemporal predictive model for multi-step forecasting task. To our knowledge, STConvS2S is the first 3D CNN-based architecture built as an end-to-end trainable model, suitable to satisfy the causal constraint and predict flexible length output sequences (i.e., not limited to be equal to the input sequence length). 

We compared STConvS2S to RNN-based architectures through experimental studies in terms of both predictive performance and time efficiency. The proposed architecture matches or outperforms state-of-the-art methods on meteorological datasets obtained from satellites and in-situ stations - CHIRPS \citep{Funk15}, and climate model - CFSR \citep{Saha14}.

The contributions of this paper are twofold. Firstly, we provide two variants of the STConvS2S architecture that satisfy the causal constraint. One adapts the causal convolution in 3D convolutional layers, and the other introduces a new approach that strategically applies a reverse function in the sequence. Secondly, we devise a temporal generator block designed to extend the length of the output sequence, which encompasses a new application of the transposed convolutional layers.

The rest of this paper is organized as follows. Section \ref{sec:works} discusses works related both to weather forecasting and spatiotemporal architectures. Section \ref{sec:problem} presents the formulation of the spatiotemporal data forecasting problem. Section \ref{sec:architecture} describes our proposed deep learning architecture. Section \ref{sec:experiments} presents our experiments and results. Section \ref{sec:conclusion} provides the conclusions of the paper.

\section{Related work} \label{sec:works}

Several statistical methods and machine learning techniques have been applied to historical data about temperature, precipitation, and other meteorological variables to predict the weather conditions. Auto-regressive integrated moving average (ARIMA) are traditional statistical methods for times series analysis \citep{Babu12}. Other studies have also applied artificial neural networks (ANN) to time series prediction in weather data, such as temperature measurements \citep{Corchado99, Baboo10, Mehdizadeh18}. Recently, some authors have been developing new approaches based on deep learning to improve time series forecasting results, in particular, using LSTM networks. Traffic flow analysis \citep{Yang19}, displacement prediction of landslide \citep{Xu18}, petroleum production \citep{Sagheer19} and sea surface temperature forecasting \citep{Zhang17} are some applications that successfully use LSTM architectures. However, these approaches (addressed to time series) are unable to capture spatial dependencies in the observations.

Spatiotemporal deep learning models deal with spatial and temporal contexts simultaneously. In \citet{Shi15}, the authors formulate weather forecasting as a sequence-to-sequence problem, where the input and output are 2D radar map sequences. In addition, they introduce the convolutional LSTM (ConvLSTM) architecture to build an end-to-end model for precipitation nowcasting. The proposed model includes the convolution operation into LSTM network to capture spatial patterns. \citet{Kim19} also define their problem as a sequence task and adopt ConvLSTM for extreme climate event forecasting. Their model uses hurricane density map sequences as spatiotemporal data. The work proposed in \citet{Souto18} implements a spatiotemporal aware ensemble approach adopting ConvLSTM architecture. Based on \citet{Shi15}, \citet{Wang17} present a new LSTM unit that memorizes spatial and temporal variations in a unified memory pool. In \citet{Wang19}, they present an improved memory function within LSTM unit adding non-stationarity modeling. Although related to the use of deep learning for climate/weather data, our model adopts only CNN rather than a hybrid approach that combines CNN and LSTM.

Some studies have applied spatiotemporal convolutions \citep{Yuan18,Tran18} for video analysis and action recognition. In \citet{Tran18}, the authors compare several spatiotemporal architectures using only 3D CNN and show that factorizing the 3D convolutional kernel into separate and successive spatial and temporal convolutions produces accuracy gains. A  limitation of both 3D CNN or factorized 3D CNN \citep{Tran18} is the lack of causal constraint, violating the temporal order. \citet{Singh19} and \citet{Cheng19} factorize the 3D convolution as \citet{Tran18}. \citet{Singh19} propose a recurrent convolution unit approach to address causal constraint in temporal learning for action recognition tasks, and \citet{Cheng19} satisfy the causal constraint by adopting causal convolution in separate and parallel spatial and temporal convolutions. We also adopt a factorized 3D CNN, but with a different implementation, where Figure \ref{fig:architecture-comparison} highlights our approach. In contrast to \citet{Singh19}, we use an entirely CNN approach, and to \citet{Cheng19}, besides not using parallel convolutions when adopting a causal convolution, we introduce a new method to not violate the temporal order (details in Section~\ref{subsec:temporal-block}).

Following the success of 2D CNN in capturing spatial correlation in images, \citet{Xu19} propose a model to predict vehicle pollution emissions using 2D CNN to capture temporal and spatial correlation separately. \citet{Racah17} use a 3D CNN in an encoder-decoder architecture for extreme climate event detection. Their architecture consists of a downsampling path in the encoder using a stack of convolutional layers, and an upsampling path in the decoder using a stack of transposed convolutional layers.  Their model adopts the typical use of transposed convolutional layers to reconstruct the output to match the entire input dimension. Instead, we use these layers to generate an output with a larger dimension, different from the dimensions of the input. Furthermore, unlike our work, they do not satisfy the causal constraint in their models.

\section{Problem Statement} \label{sec:problem}

Spatiotemporal data forecasting can be modeled as a sequence-to-sequence problem. Thus, the observations of spatiotemporal data (e.g. meteorological variables) measured in a specific geographic region over a period of time serve as the input sequence to the forecasting task.  More formally, we define a spatiotemporal dataset as $[\widetilde{X}^{(1)}, \widetilde{X}^{(2)},\ldots, \widetilde{X}^{(m)}]$ with $m$ samples of $\widetilde{X}^{(i)} \in \mathbb{R}^{T \times H \times W \times C}$, where $1 \leq i \leq m$. Each training example is a tensor $\widetilde{X}^{(i)} = [X_1^{(i)}, X_2^{(i)},\ldots,X_T^{(i)}]$, that is a sequence of $T$ observations containing historical measurements. Each observation $X_j^{(i)} \in \mathbb{R}^{H \times W \times C}$, for $j = 1, 2,\ldots, T$  (i.e. the length of input sequence), consists of a $H \times W$ grid map that determines the spatial location of the measurements, where $H$ and $W$ represent the latitude and longitude, respectively. In the observations, $C$ represents how many meteorological variables (e.g. temperature, humidity) are used simultaneously in the model. This structure is analogous to 2D images, where $C$ would indicate the amount of color components (RGB or grayscale).

Modeled as sequence-to-sequence problem in Equation \ref{eq:prediction}, the goal of spatiotemporal data forecasting is to apply a function $f$ that maps an input sequence of past observations, satisfying the causal constraint at each time step $t$, in order to predict a target sequence $\hat{X} \in \mathbb{R}^{H \times W \times C}$, where the length $T^{\prime\prime}$ of output sequence may differ from the length $T$ of input sequence.

\begin{equation}
     \hat{X}_{t+1},\hat{X}_{t+2},\ldots, \hat{X}_{t+T^{\prime\prime}} = f(X_{t-T+1},\ldots,X_{t-1},X_{t}) 
\label{eq:prediction}    
\end{equation}

\section{STConvS2S architecture} \label{sec:architecture}

\begin{figure*}
    \includegraphics[width=\linewidth]{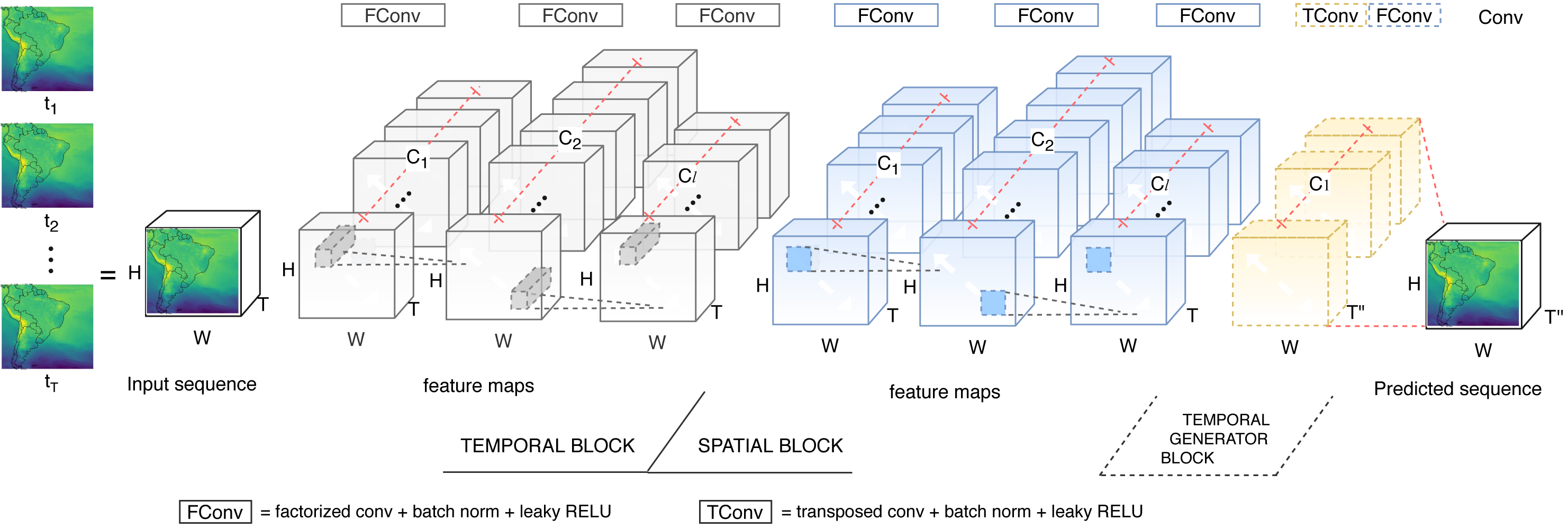}
    \caption{An illustration of STConvS2S architecture, which comprises three components: temporal block, spatial block, and temporal generator block. Each block is a set of layers. The temporal block learns a temporal representation of the input sequence, the spatial block extracts spatial features from the output of the previous block. On top of the spatial block, there is the temporal generator block designed to increase the sequence length  $T$ if the task requires a longer predictive horizon, where $T^{\prime \prime} \geqslant T$. Finally, the output of this block is further fed into a final convolutional layer to complete the prediction}
    \label{fig:abstraction-seq2seq}
\end{figure*}

STConvS2S is an end-to-end deep neural network suited for learning spatiotemporal predictive patterns, which are common in domains weather forecasting. Our approach makes multi-step (sequences) prediction without feeding the predicted output back into the input sequence. Figure \ref{fig:abstraction-seq2seq} is an overview of our proposed deep learning architecture.

Although some methods for weather forecasting using a radar echo dataset apply a hybrid approach, combining 2D CNN (to learn spatial representations) and LSTM (to learn temporal representations) \citep{Shi15, Wang17, Wang19}, our method uses only 3D convolutional layers to learn spatial and temporal contexts. Distinct from conventional convolution applied in some 3D CNN architectures \citep{Tran15, Tran18, Racah17}, during temporal learning, STConvS2S takes care not to depend on future information, a crucial constraint on forecasting tasks. Another core feature of our designed network is the capability to allow flexible output sequence length, which means the possibility to predict many time-steps ahead, regardless of the fixed-length of the input sequence. In the following, we provide more details about the components which comprise our architecture.

\subsection{Factorized 3D convolutions}

\begin{figure}
    \begin{minipage}[b]{0.33\columnwidth}
    \centering
    \includegraphics{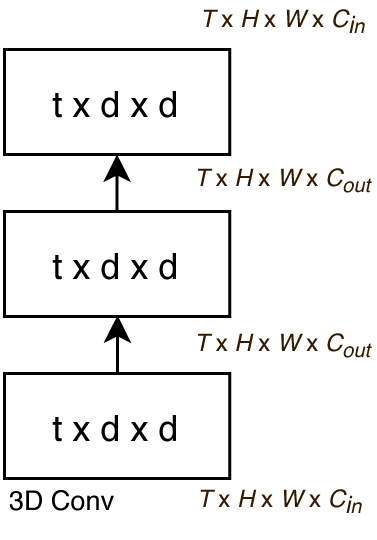}\\
    \subcaption{}
    \end{minipage}%
    \begin{minipage}[b]{0.33\columnwidth}
    \centering
    \includegraphics{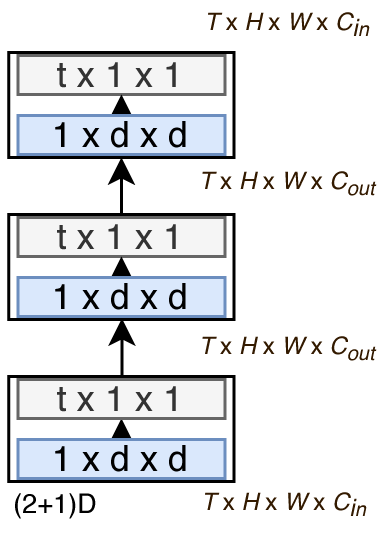}\\
    \subcaption{}
    \end{minipage}%
    \begin{minipage}[b]{0.33\columnwidth}
    \centering
    \includegraphics{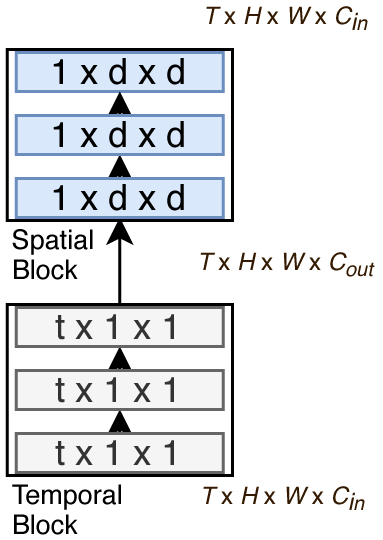}\\
    \subcaption{}
    \end{minipage}%
    \caption{Comparison of convolution operations applied in three convolutional layers. The spatial kernel is defined as $1 \times d \times d$ and the temporal kernel as $t \times 1 \times 1$, where $d$ and $t$ are the kernel size in spatial ($H \times W$) and time ($T$) dimensions, respectively.  (a) Representation of the standard 3D convolution operation using the $t \times d \times d$ kernel. (b) Factorized 3D kernels proposed in \citet{Tran18} as successive spatial and temporal convolution operations in a unique block called (2+1)D. (c) Our proposal for the factorized 3D kernels usage is in separate blocks. First, the temporal block stacks three convolutional layers, each performing convolutions using only the temporal kernel. Likewise, the spatial block applies the spatial kernel to its layers.}
    \label{fig:architecture-comparison}
\end{figure}

Instead of adopting a conventional $t \times d \times d$ kernel for 3D convolutional layers, where $d$ and $t$ are the kernel size in space ($H \times W$) and time ($T$) dimensions, respectively, we use a factorized 3D kernel adapted from R(2+1)D network, proposed in \citet{Tran18}. The factorized kernel $1 \times d \times d$ and $t \times 1 \times 1$ split the convolution operation of one layer into two successive operations, named as a spatial convolution and a temporal convolution in their work. In our new architecture, we take a different approach: operations are not successive inside each convolutional layer. Instead, the factorized kernels are separated into two blocks, giving them specific learning skills. The temporal block applies the $t \times 1 \times 1$ kernel in its layers to learn only temporal dependencies, while the next component, the spatial block, encapsulates spatial dependencies using $1 \times d \times d$ kernel. Figure \ref{fig:architecture-comparison} schematically illustrates the difference between these three approaches.

Compared to the full 3D kernel applied in standard convolutions, the kernel decomposition used in STConvS2S offers the advantage of increasing the number of nonlinearities in the network (additional activation functions between factorized convolutions), which leads to an increase in the complexity of representable patterns \citep{Tran18}. An advantage of our proposed approach over the (2+1)D block is flexibility since temporal and spatial blocks can have a distinct number of layers, facilitating their optimization.

\subsection{Temporal Block} \label{subsec:temporal-block}

In STConvS2S, the temporal block is a stack of 3D convolutional layers which adopt $t \times 1 \times 1$ kernel during convolutions. Each layer receives a 4D tensor with dimensions $T \times H \times W \times C_{l-1}$ as input, where $C_{l-1}$ is the number of filters used in the previous layer ($l-1$), $T$ is the  sequence length (time dimension), $H$ and $W$ represent the size of the spatial coverage for latitude and longitude, respectively. Within the block, the filters $C$ are increased twice in the feature maps as the number of layers increases, but the final layer reduces them again to the number of filters initially defined. 

In detail, this block uses batch normalization and leaky rectified linear unit (LeakyReLU) with a negative slope set as 0.01 after each convolutional layer. This block discovers patterns over the time dimension $T$ exclusively. Besides, since we are using 3D convolutional layers to analyze historical series of events, we must prevent data leakage from happening. That is, the model should not violate the temporal order and should ensure that, at step $t$, the learning process uses no future information from step $t+1$ onward. To satisfies this constraint, we propose two variants of the temporal block.

\begin{figure}[h]
    \centering
    \includegraphics[width=0.5\linewidth]{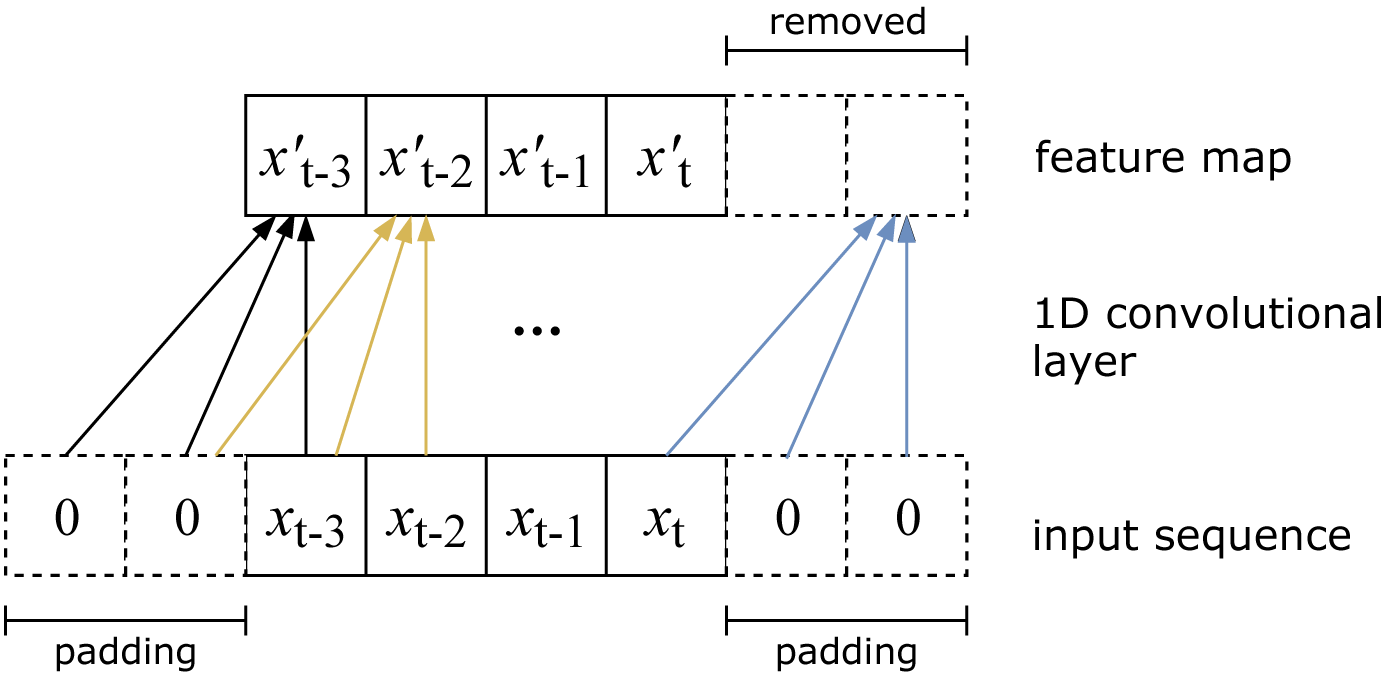}
    \caption{Causal convolution operation in a 1D convolutional layer (used to simplify the illustration) with $k = 3$ (kernel size). Input is padded by $k-1$ elements to avoid learning future information. To ensure that the output feature map has the same length as the input, the last $k-1$ elements are removed since they are related to the zeros added to the right of the input.}
    \label{fig:causal-conv}
\end{figure}

\textbf{Temporal Causal Block.} We name our architecture as \emph{STConvS2S-C} when it adopts this block to learn the temporal patterns. We apply causal convolutions within the block to incorporate the ability to respect the temporal order during learning in convolutional layers. Causal convolution was originally presented in WaveNet \citep{Oord16} for 1D CNN and applied with factorized 3D convolutions in \citet{Cheng19}. This technique can be implemented by padding the input by $k-1$ elements, where $k$ is the kernel size. Figure \ref{fig:causal-conv} shows the operation in details.

\textbf{Temporal Reversed Block.} When dealing with historical data, respecting the temporal order (causal constraint) is an essential behavior of deep learning models. This because, in real applications, future information is not available in forecasting. The common approach in the literature for adapting convolutional layers to satisfy this constraint is through causal convolutions. Here, we introduce a better alternative to avoid violating the temporal order, applying a function $\psi$ in the time dimension to reverse the sequence order. This function is a linear transformation $\psi:\mathbb{R}^{T \times H \times W \times C} \rightarrow \mathbb{R}^{T \times H \times W \times C}$. The architecture is named as \emph{STConvS2S-R} when composed with this block. Formally, STConvS2S-R computes the output feature map $R$ of a temporal reversed block using

\begin{equation}
     R^{\prime}_{1:l_r} = 
    \begin{cases}
      g(W_u \ast \psi(I_u) + b_u) , & \text{if}\ u = 1 \\
      g(W_u \ast I_u + b_u), & \text{if}\ 2 \leqslant u  \leqslant l_r
    \end{cases}
\label{eq:inner-output}    
\end{equation}

\begin{equation}
     R = \psi(R^{\prime}_{l_r})
\label{eq:output}    
\end{equation}

where $W_{1:l_r}$ and $b_{1:l_r}$ is the learnable weight tensor and bias term in $l_r$ layers of this block, $\ast$ denotes a convolution operator and $g(\cdot)$ is a non-linear activation function. For the first layer of the temporal reversed block, $I_1$ is the input sequence $\widetilde{X}$ previously defined in Section \ref{sec:problem}, and for the subsequent layers,  $I_{2:l_r}$ is the feature map calculated in the previous layer $R^{\prime}_{l_{r}-1}$.

\subsection{Spatial Block}

The spatial block is built on top of the temporal block and has a similar structure with batch normalization and LeakyReLU as non-linearity. In contrast, each 3D convolutional layer of this block extracts only spatial representations since kernel decomposition allows us to analyze the spatial and temporal contexts separately. 
In the STConvS2S, each feature map generated has a fixed-length in $H \times W$ dimensions and, to ensure this, the input in the spatial block is padded following $p = \frac{k_s - 1}{2}$, where $k_s$ is the size of spatial kernel. This design choice differentiates our model from 3D encoder-decoder architecture \citep{Racah17}, which needs to stack upsample layers after all convolutional layers, due to the downsampling done in the latter.

\subsection{Temporal Generator Block}

In addition to ensuring that our model satisfies the causal constraint, another contribution of our work is generating output sequences with longer lengths than the length of the input sequence. When CNNs are used for sequence-to-sequence learning, such as multi-step forecasting, the length of the output sequence must be the same size or shorter than the input sequence \citep{Gehring17, Bai18}. To tackle this limitation, we designed a component placed on top of the spatial block used when the task requires a more extended sequence (e.g., from the previous 5 grids, predict the next 15 grids). First, we compute the intermediate feature map $G^{\prime}$:

\begin{equation}
     G^{\prime}_{1:l_{g_t}} = tconv(I_{1:l_{g_t}})
\label{eq:inner-output} 
\end{equation}
where $l_{g_t} = \ceil*{\frac{T^{\prime \prime} - T}{2 T}}$ is the number of transposed convolutional layers ($tconv$) necessary to guarantee that $G^{\prime}$ has the size of time dimension $T_{G^{\prime}} \geqslant T^{\prime \prime}- T$. The kernel size, stride and padding of $tconv$ are fixed and extend the feature map by a factor of 2 in time dimension only. For the first layer, $I_1$ is the output of the spatial block $S$ and for the other layers, $I_{2:l_{g_t}}$ is the feature map calculated in the previous layer $G^{\prime}_{l_{g_t}-1}$. Follow, given $G^{\prime}$ and $S$, we can compute $G^{\prime \prime}$:

\begin{equation}
     G^{\prime \prime} = \rho(S \oplus G^{\prime})
\label{eq:g-prime-prime}    
\end{equation}

In the equation above, $\oplus$ denotes a concatenation operator in the time dimension and $\rho(\cdot)$ is a function to ensure that the feature map $G^{\prime \prime}$ matches exactly the length $T^{\prime \prime}$ of the desired output sequence. Finally, the output feature map $G$ of this block can be defined as

\begin{equation}
     G_{1:l_{g_c}} = g(W_{1:l_{g_c}} \ast I_{1:l_{g_c}}  + b_{1:l_{g_c}})
\label{eq:output}    
\end{equation}

where $l_{g_c} = \floor*{\frac{T^{\prime \prime}}{T}}$ is the number of convolutional layers that use factorized kernels as in the spatial block. $W_{1:l_{g_c}}$ and $b_{1:l_{g_c}}$ is the learnable weight tensor and bias term in $l_{g_c}$ layers, $\ast$ denotes a convolution operator and $g(\cdot)$ is a non-linear activation function. For the first convolutional layer, $I_1$ is $G^{\prime \prime}$. Unlike the temporal and spatial block, where the number of layers is a defined hyperparameter to execute the model, in the temporal generator block $l_{g_t}$ and $l_{g_c}$ are calculated based on the length $T^{\prime \prime}$ of the desired output sequence and the length $T$ of the input sequence (the size of time dimension).

\section{Experiments} \label{sec:experiments}
We perform experiments on two publicly available meteorological datasets containing air temperature and precipitation values to validate our proposed architecture. The deep learning experiments were conducted on a server with a single Nvidia GeForce GTX1080Ti GPU with 11GB memory. We executed the ARIMA methods on 8 Intel i7 CPUs with 4 cores and 66GB RAM. We start by explaining the datasets (Section~\ref{sec:datasets}) and evaluation metrics (Section~\ref{sec:Metrics}). Further, we describe the main results of the experiments for each dataset (Section~\ref{sec:cfsr-results} and \ref{sec:chirps-results}) and summarize the results of ablation studies (Section~\ref{sec:ablation}). 

\subsection{Datasets} 
\label{sec:datasets}

The CFSR\footnote{\url{https://climatedataguide.ucar.edu/climate-data/climate-forecast-system-reanalysis-cfsr}} is a reanalysis\footnote{Scientific method used to produce best estimates (analyses) of how the weather is changing over time \citep{Fujiwara17}.} product that contains high-resolution global land and ocean data \citep{Saha14}. The data contain a spatial coordinate (latitude and longitude), a spatial resolution of 0.5 degrees (i.e., $0.5^{\circ} \times 0.5^{\circ}$ area for each grid cell) and a frequency of 6 hours for some meteorological variables, such as air temperature and wind speed.

\begin{figure}
    \centering
    \begin{minipage}[b]{0.35\linewidth}
        \centering
        \includegraphics[width=\textwidth]{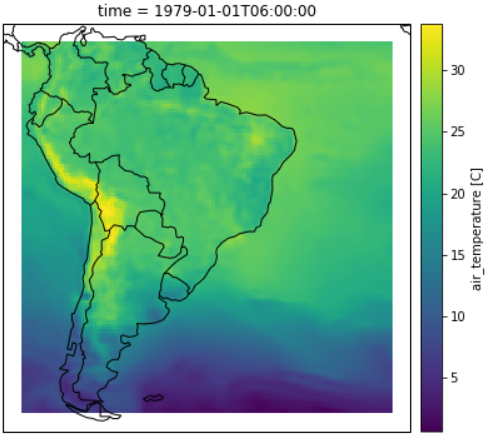}
        \subcaption{CFSR-temperature dataset}
    \end{minipage}
    \hspace{0.5cm}
    \begin{minipage}[b]{0.35\linewidth}
        \centering
        \includegraphics[width=\textwidth]{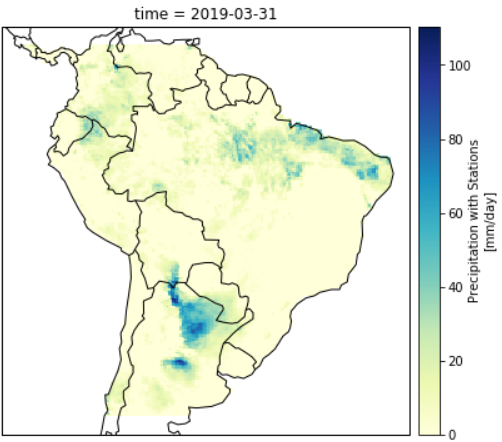}
        \subcaption{CHIRPS-rainfall dataset}
    \end{minipage}
    \caption{Spatial coverage of the datasets used in all experiments. (a) It shows the selected grid on January 1, 1979 with air temperature values.(b) It shows the selected grid of the sequence on March 31, 2019 with rainfall values.}
    \label{fig:datasets}
\end{figure}

In the experiments, we use a subset of CFSR with the air temperature observations from January 1979 to December 2015, covering the space in 8$^{\circ}$N-54$^{\circ}$S and 80$^{\circ}$W-25$^{\circ}$W as shown in Figure \ref{fig:datasets} (a). As data preprocessing, we scale down the grid to $32 \times 32$ in the $H$ and $W$ dimensions to fit the data in GPU memory. The other dataset, CHIRPS\footnote{\url{https://chc.ucsb.edu/data/chirps}}, incorporates satellite imagery and in-situ station data to create gridded rainfall times series with daily frequency and spatial resolution of 0.05 degrees \citep{Funk15}. We use a subset with observations from January 1981 to March 2019 and apply interpolation to reduce the grid size to $50 \times 50$. Figure \ref{fig:datasets} (b) illustrates the coverage space 10$^{\circ}$N-39$^{\circ}$S and 84$^{\circ}$W-35$^{\circ}$W adopted in our experiments.

Similar to \citet{Shi15}, we define the input sequence length as 5, which indicates the use of the previous five grids to predict the next $T''$ grids. Thus, the input data shapes to the deep learning architectures are $5 \times 32 \times 32 \times 1$  for CFSR dataset and $5 \times 50 \times 50 \times 1$ for CHIRPS dataset. The value 1 in both dataset shapes indicates the one-channel (in this aspect similar to a grayscale image), 5 is the size of the sequence considered in the forecasting task, and 32 and 50 represent the numbers of latitudes and longitudes used to build the spatial grid in each dataset. 

We create 54,041 and 13,960 grid sequences from the temperature dataset and rainfall datasets, respectively. Finally, we divide both datasets into non-overlapping training, validation, and test set following 60\%, 20\%, and 20\% ratio, in this order. The adoption of temperature and rainfall datasets in our experimental evaluation relies on the fact that they are the two main meteorological variables. Research about their spatiotemporal representation is relevant to short-term forecasting and improves the understanding of long-term climate variability \citep{Rahman17}. However, the proposed architecture is suitable for other meteorological variables or other domains, as long as the training data can be structured as defined in Section \ref{sec:problem}.

\subsection{Evaluation metrics}
\label{sec:Metrics}
To evaluate the proposed architecture, we compare our results against ARIMA models, traditional statistical approaches for time series forecasting, and state-of-the-art models for spatiotemporal forecasting. To accomplish this, we use the two evaluation metrics presented in Equation \ref{eq:rmse} and \ref{eq:mae}. 

RMSE, denoted as $E_r$, is based on MSE metric, which is the average of squared differences between real observation and prediction. The MSE square root gives the results in the original unit of the output, and is expressed at a specific spatiotemporal volume as:

\begin{equation}
E_r(T,H,W) = \sqrt{\frac{1}{N} \sum_{n=1}^{N} \sum_{t \in T} \sum_{h \in H} \sum_{w \in W} [x(t,h,w) - \hat{x}(t,h,w)]^2}
\label{eq:rmse}
\end{equation}
where $N$ is the number of test samples, $x(t,h,w)$ and $\hat{x}(t,h,w)$ are the real and predicted values at the location $h$ and $w$ at time $t$, respectively. 

MAE, denoted as $E_m$, is the average of differences between real observation and prediction, which measures the magnitude of the errors in prediction. MAE also provides the result in the original unit of the output, and is expressed at a specific spatiotemporal volume as:
\begin{equation}
E_m(T,H,W) = \frac{1}{N} \sum_{n=1}^{N} \sum_{t \in T} \sum_{h \in H} \sum_{w \in W} |x(t,h,w) - \hat{x}(t,h,w)|
\label{eq:mae}
\end{equation}
where $N$, $t$, $h$, $w$ are defined as shown in Equation \ref{eq:rmse}.

\subsection{CFSR Dataset: results and analysis}
\label{sec:cfsr-results}

We first conduct experiments with distinct numbers of layers, filters, and kernel sizes to investigate the best hyperparameters to fit the deep learning models. As a starting point, we set the version $1$ based on the settings described in \citet{Shi15} with two layers, each containing 64 filters and a kernel size of 3\footnote{$3 \times 3$ kernel for ConvLSTM, PredRNN, and MIM. $3 \times 1 \times 1$ temporal kernel and $1 \times 3 \times 3$ spatial kernel for STConvS2S.}. To make fair comparisons using the chosen datasets, we explored variations of the hyperparameters for our architecture (STConvS2S-C and STConvS2S-R) and the following state-of-the-art methods: ConvLSTM \citep{Shi15}, PredRNN \citep{Wang17}, and MIM \citep{Wang19}. Thus, for versions $2$-$4$, we defined the number of layers ($L$), kernel size ($K$) and the number of filters ($F$) in a way that would help us understand the behavior of the models during the learning process by increasing L (versions $1$ and $3$), K (versions $2$ and $4$) or F (versions $2$ and $3$). In the training phase, we perform for all models mini-batch learning with 50 epochs, and RMSprop optimizer with a learning rate of $10^{-3}$.

\begin{figure}
    \centering
    \begin{minipage}[b]{0.47\linewidth}
        \centering
        \includegraphics[width=\textwidth]{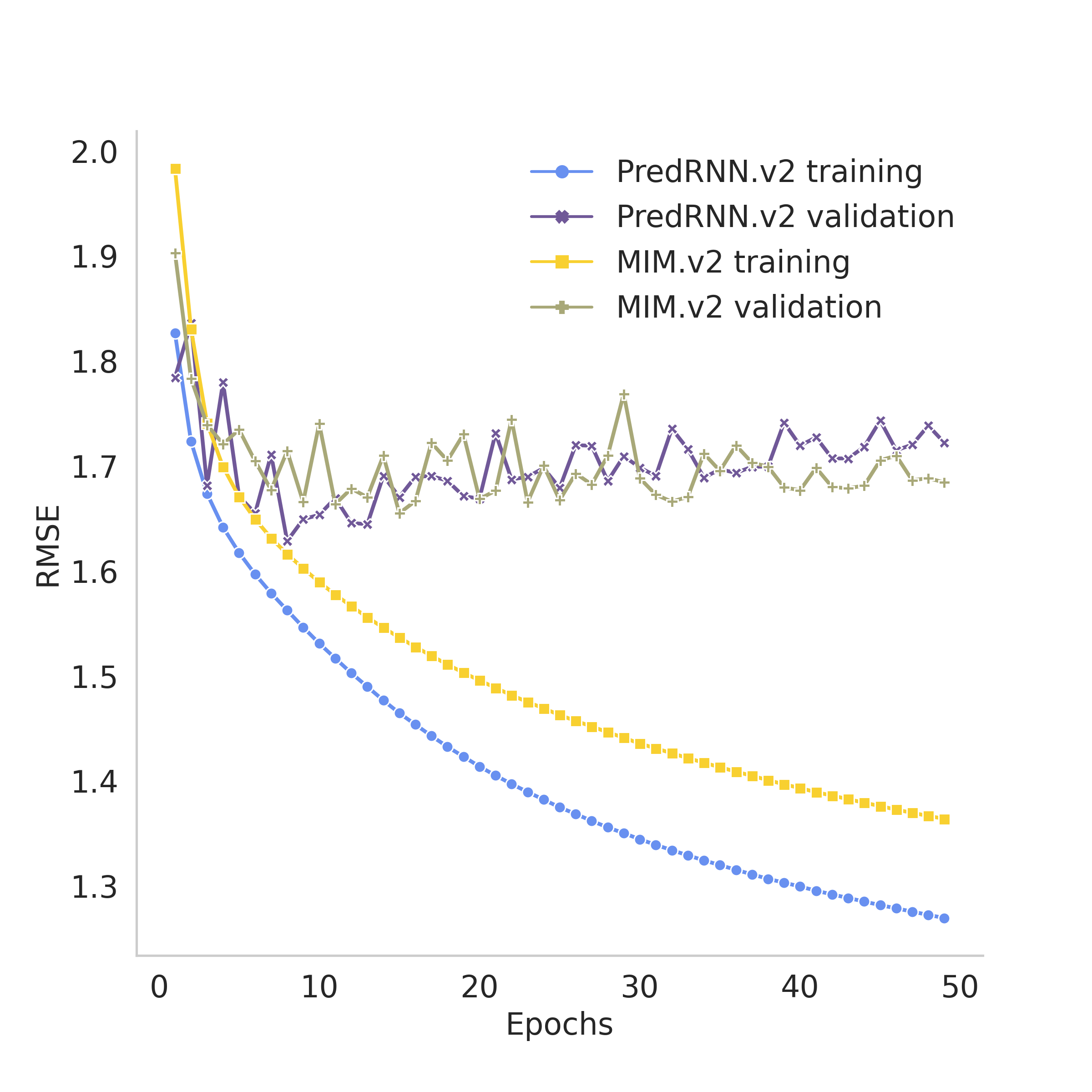}
        \subcaption{Learning curve - overfitting}
    \end{minipage}
    \begin{minipage}[b]{0.47\linewidth}
        \centering
        \includegraphics[width=\textwidth]{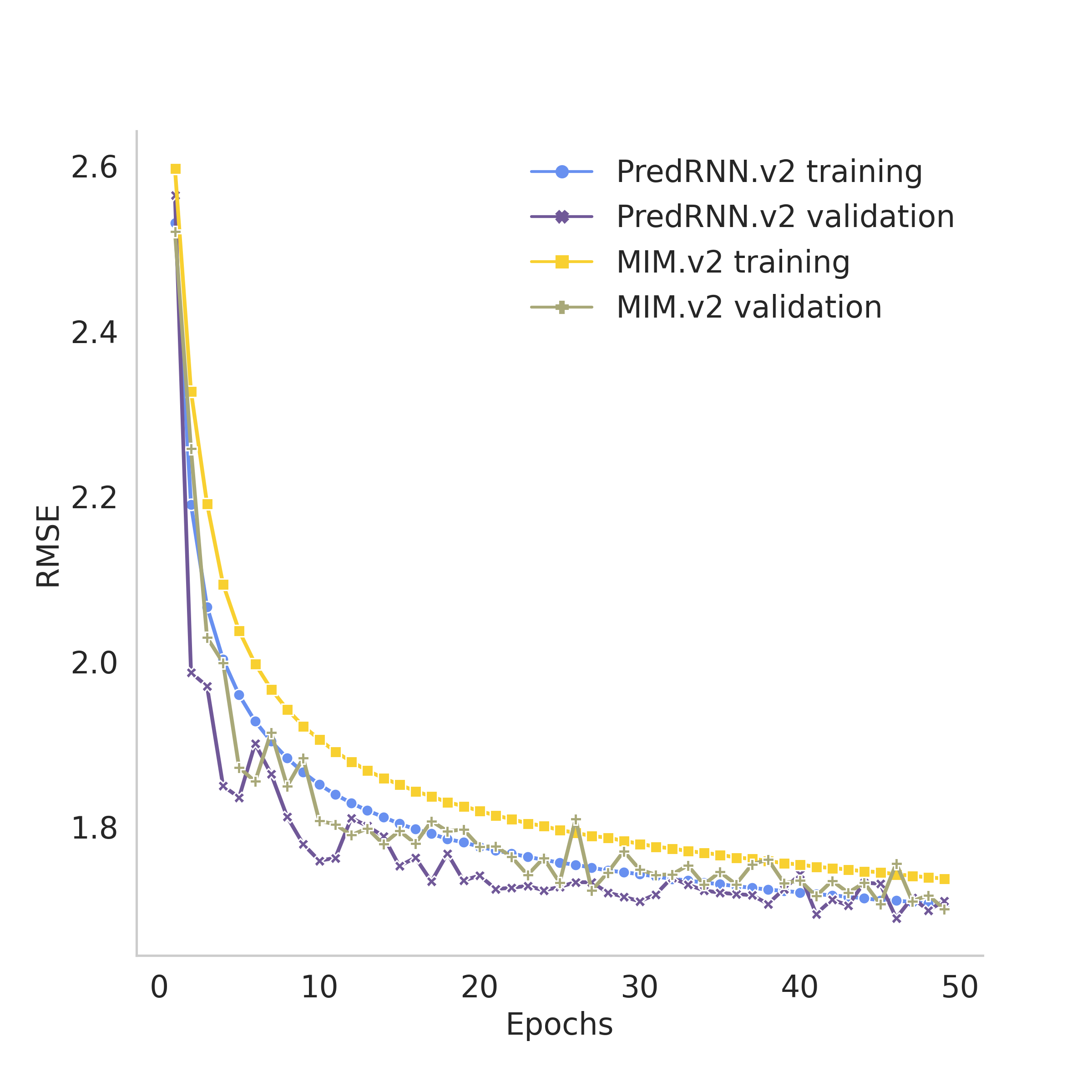}
        \subcaption{Learning curve - dropout 0.5}
    \end{minipage}
        \begin{minipage}[b]{0.5\linewidth}
        \centering
        \includegraphics[width=\textwidth]{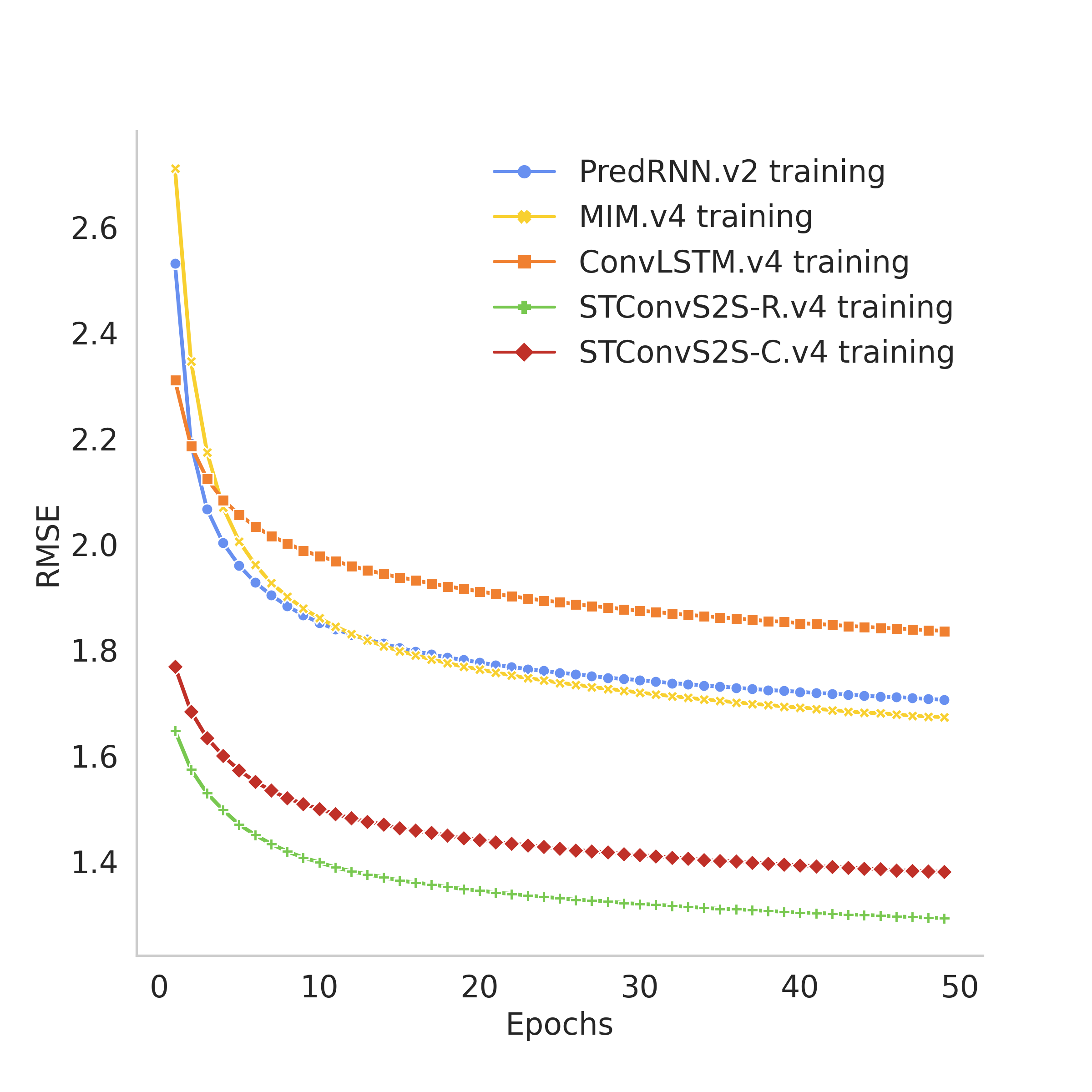}
        \subcaption{Training curve for all models}
    \end{minipage}
    \caption{Learning curves after running 50 epochs on temperature dataset (CFSR). (a) To exemplify, we select version $2$ to illustrate the overfitting observed when analyzing the training and validation curve of PredRNN and MIM models. (b) The same version and models using dropout to improve its generalization. (c) Comparison of training curve for the best version for each model. Our models (STConvS2S-R and STConvS2S-C) achieved lower RMSE and thus, better ability to learn spatiotemporal representations.}
    \label{fig:lineplot}
\end{figure}

We applied dropout after convolutional layers during the training of PredRNN and MIM models \footnote{STConvS2S and ConvLSTM models do not overfit during training on any version.} to reduce the model complexity and avoid overfitting. Without dropout, these models do not generalize well for this dataset and make less accurate predictions in the validation set. We adopt 0.5 as the dropout rate for both models after evaluating the best rate employing the grid search technique, which performed several experiments changing the rate by \{0.3, 0.5, 0.8\}. Figure \ref{fig:lineplot} (a) and (b) illustrate the differences in the learning curve, where the former shows a high error on the validation set early in the training stage for both models, and the latter indicates the learning curve with dropout applied.

As a sequence-to-sequence task, we use the previous five grids, as we established before in Section \ref{sec:datasets}, to predict the next five grids (denoted as {$5 \rightarrow 5$}). Table \ref{tab:cfsr-exp1} provides the models considered in our investigation with four different settings, the values of the RMSE metric on the test set, the training time, and the memory usage by GPU. The results present the superiority of version $4$, which has the highest values of L and K, reaching the lowest RMSE for all models, except PredRNN, where version $2$ is superior. Another aspect to note is that when increasing the number of filters (versions $2$ and $3$) the impact is more significant in the training time than when increasing the number of layers  (versions $1$ and $3$) for state-of-the-art models, indicating that version $2$ is faster for these RNN-based architectures. This impact is not seen in our models (CNN-based) compared to version $3$, as in versions $1$ and $2$ they spend almost the same time during training, showing that STConvS2S models have more stability when increasing the hyperparameters.

\begin{figure}
    \centering
    \includegraphics[width=0.70\linewidth]{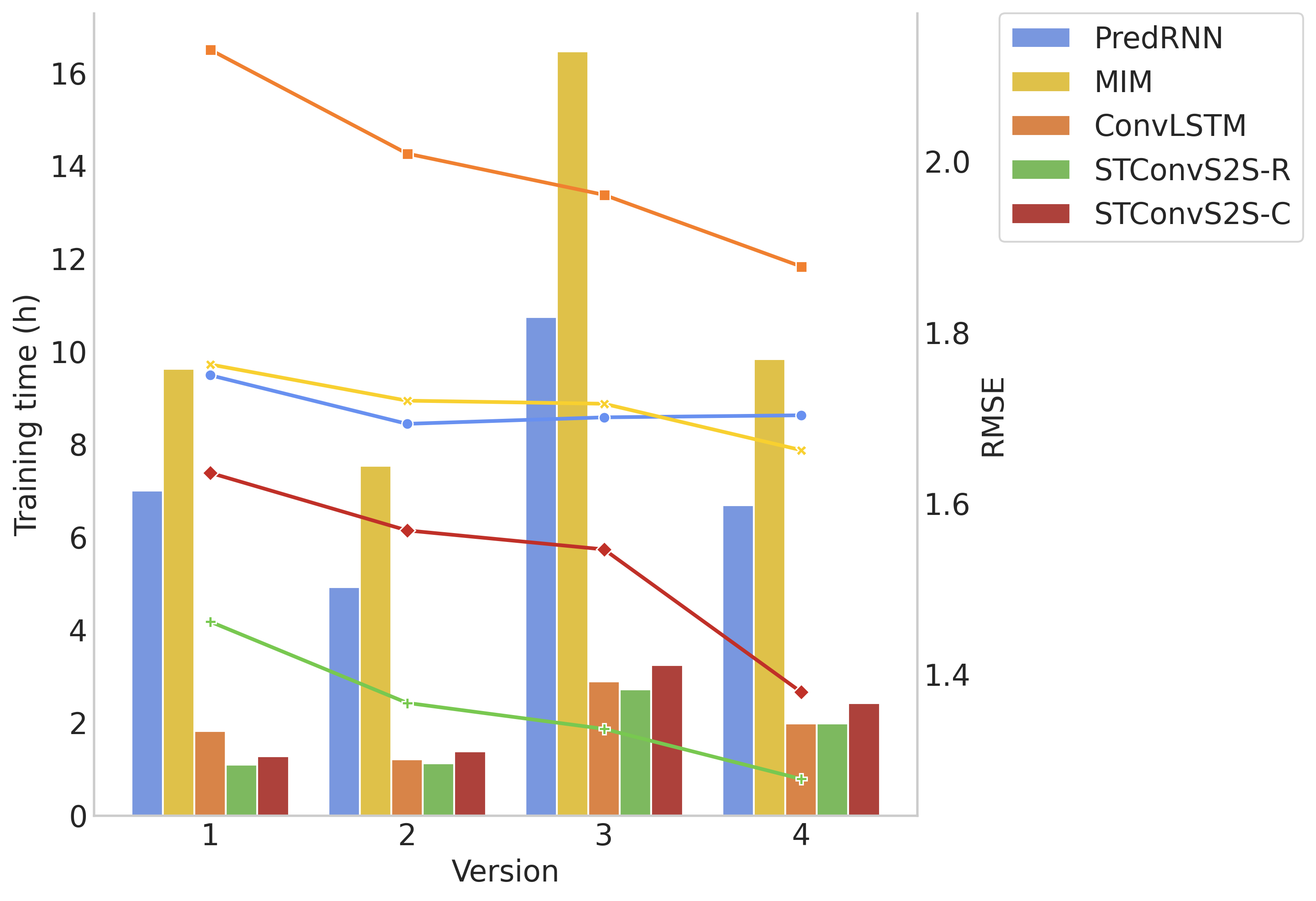}
    \caption{Comparison between training time in hours (bar plot) and RMSE (line plot) for each model version on temperature dataset (CFSR).}
    \label{fig:rmse-all-models}
\end{figure}

\begin{table}
\centering
\caption{Evaluation of different settings on the CFSR dataset for STConvS2S and state-of-the-art methods, where the best version has the lowest RMSE value.}
\label{tab:cfsr-exp1}
\begin{tabular}{lccccc@{\extracolsep{\fill}}} 
\toprule
                                  &         &                  & \multicolumn{3}{c}{$5 \rightarrow 5$}                         \\
\cmidrule{4-6}                            
Model         & Version & Setting   & RMSE              & Training time     &  \parbox{2.5cm}{\centering Memory usage (MB)}      \\ 
\midrule
\multirow{4}{*}{\parbox{3cm}{ConvLSTM \\ \citep{Shi15}}} 
    & 1       & L=2, K=3, F=64    & 2.1306              & 01:49:16           & 1119                \\
    & 2       & L=3, K=3, F=32    & 2.0090              & \textbf{01:12:16}  & \textbf{920}        \\ 
    & 3       & L=3, K=3, F=64    & 1.9607              & 02:53:00           & 1358                \\
    & 4       & L=3, K=5, F=32    & \textbf{1.8770}     & 01:58:52           & 922      \\

\midrule
\multirow{4}{*}{\parbox{3cm}{PredRNN \\ \citep{Wang17}}}      
    & 1       & L=2, K=3, F=64    & 1.7497              & 06:59:45           & 3696                \\
    & 2       & L=3, K=3, F=32    & \textbf{1.6928}     & \textbf{04:55:19}  & \textbf{2880}       \\ 
    & 3       & L=3, K=3, F=64    & 1.7004              & 10:44:25           & 5242                \\
    & 4       & L=3, K=5, F=32    & 1.7028              & 06:41:14           & 2892       \\
\midrule
\multirow{4}{*}{\parbox{3cm}{MIM \\ \citep{Wang19}}}      
    & 1       & L=2, K=3, F=64    & 1.7623              & 09:37:07           & 4826                \\
    & 2       & L=3, K=3, F=32    & 1.7199              & \textbf{07:31:59}  & \textbf{4124}       \\ 
    & 3       & L=3, K=3, F=64    & 1.7163              & 16:27:42           & 7789                \\
    & 4       & L=3, K=5, F=32    & \textbf{1.6621}     & 09:49:40           & 4145                \\
\midrule
\multirow{4}{*}{\parbox{3cm}{STConvS2S-C (ours)}}      
    & 1       & L=2, K=3, F=64    & 1.6355              & \textbf{01:16:40}  & \textbf{991}        \\
    & 2       & L=3, K=3, F=32    & 1.5681              & 01:22:42           & 1021                \\ 
    & 3       & L=3, K=3, F=64    & 1.5459              & 03:14:25           & 1554                \\
    & 4       & L=3, K=5, F=32    & \textbf{1.3791}     & 02:25:26           & 1040                 \\
\midrule
\multirow{4}{*}{\parbox{3cm}{STConvS2S-R (ours)}}      
    & 1       & L=2, K=3, F=64    & 1.4614              & \textbf{01:05:48}  & \textbf{880}        \\
    & 2       & L=3, K=3, F=32    & 1.3663              & 01:07:33           & 891                 \\ 
    & 3       & L=3, K=3, F=64    & 1.3359              & 02:42:53           & 1283                \\
    & 4       & L=3, K=5, F=32    & \textbf{1.2773}     & 01:58:39           & 895                 \\
\bottomrule
\end{tabular}
\end{table}

To improve the comprehension of the analysis, Figure \ref{fig:rmse-all-models} highlights the differences between the performances of RMSE metric and training time for the models. As shown, STConvS2S-R and STConvS2S-C models perform favorably against the state-of-the-art models for the CFSR dataset in all versions, demonstrating that our architectures can simultaneously capture spatial and temporal correlations. Comparing the best version of each model,  our models significantly outperform the state-of-the-art architectures for spatiotemporal forecasting. In detail, STConvS2S-R (version $4$) takes only 1/4 memory space, is 5x faster in training, and has achieved a 23\% improvement in RMSE over MIM (version $4$), the RNN-based model with better performance. These results reinforce that our models have fewer parameters to optimize than MIM and PredRNN. Furthermore, our model can be completely parallelized, speeding up the learning process, as the output of the convolutional layers does not depend on the calculations of the previous step, as occurs in recurrent architectures. Figure \ref{fig:lineplot} (c) illustrates that STConvS2S-R has a lower training error in 50 epochs compared to other models, including STConvS2S-C, proving to be a better alternative to make CNN-based models respect the temporal order.

To further evaluate our models, we chose the most efficient version for each model to perform new experiments. For STConvS2S-R, STConvS2S-C, ConvLSTM and MIM models, version $4$ was chosen with 3 layers, 32 filters, and kernel size of 5, and for PredRNN, version $2$ with the same number of layers and filters, but with kernel size of 3. We also included a comparison with ARIMA methods to serve as baseline for deep learning models, since they are a traditional approach to time series forecasting. The experiment for the baseline takes into account the same temporal pattern and spatial coverage. Thus, predictions were performed throughout all the 1,024 time series, considering in each analysis the previous 5 values in the sequence. 

In this phase, we have not defined a specific number of epochs for each deep learning model's execution. Therefore to avoid overfitting during the training of models, we apply the early stopping technique with patience hyperparameter set to 16 on the validation dataset. As the models run with different numbers of epochs, we include the training time/epoch to be able to compare the time efficiency of the models. We train and evaluate each deep learning model 3 times and compute the mean and the standard deviation of RMSE and MAE metrics on the test set. This time, we evaluate the models in two horizons: 5-steps ({$5 \rightarrow 5$}) and 15-steps ({$5 \rightarrow 15$}). These experiments are relevant to test the capability of our model to predict a long sequence.

\begin{figure}
    \centering
    \begin{minipage}[b]{1\linewidth}
        \centering
        \includegraphics[width=0.7\textwidth]{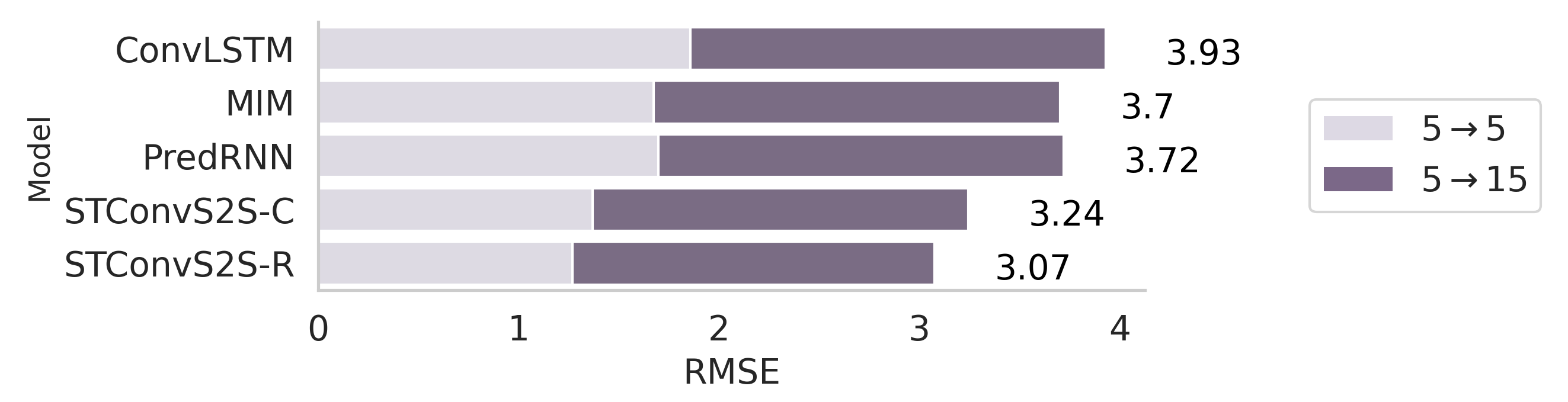}
    \end{minipage}
    \hspace{0.5cm}
    \begin{minipage}[b]{1\linewidth}
        \centering
        \includegraphics[width=0.7\textwidth]{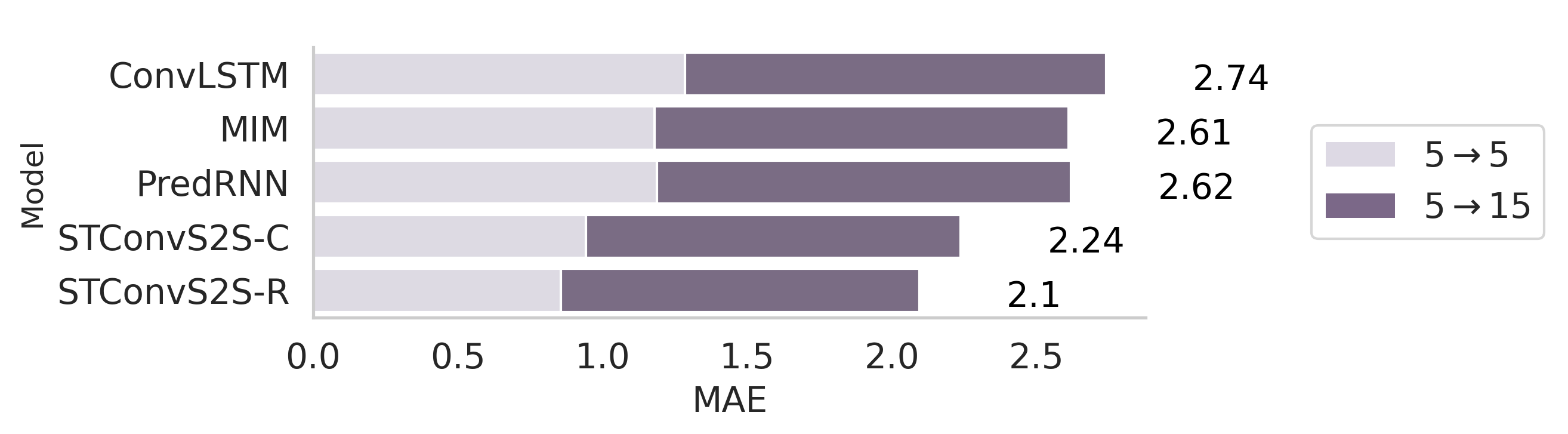}
    \end{minipage}
    \caption{Cumulative error based on both horizons ($5 \rightarrow 5$ and $5 \rightarrow 15$) using temperature dataset (CFSR). Evaluations on RMSE and MAE metrics.}
    \label{fig:stackedplot-cfsr}
\end{figure}

As shown in Table \ref{tab:dataset1}, STConvS2S-R and STConvS2S-C perform much better than the baseline on RMSE and MAE metrics, indicating the importance of spatial dependence on geoscience data since ARIMA models only analyze temporal relationships. They also outperform the state-of-the-art models in these evaluation metrics in both horizons, demonstrating that our models can be efficiently adopted to predict future observations. However, beyond that, the designed temporal generator block in STConvS2S architecture can convincingly generate a more extended sequence regardless of the fixed-input sequence length. In a closer look at the best CNN-based architecture and RNN-based architecture in the task {$5 \rightarrow 15$}, STConvS2S-R takes less memory space and is faster than PredRNN. To provide an overview, Figure \ref{fig:stackedplot-cfsr} illustrates the cumulative error based on both horizons.

\begin{table}
\centering
\caption{Performance results for temperature forecasting using the previous five observations (grids) to predict the next five observations ($5 \rightarrow 5$), and the next 15 observations ($5 \rightarrow 15$).}
\label{tab:dataset1}
\begin{adjustbox}{width=\textwidth}
\begin{tabular}{lccccc@{\extracolsep{\fill}}} 
\toprule
                  & \multicolumn{5}{c}{$5 \rightarrow 5$}      \\
\cmidrule{2-6}   
Model       & RMSE    & MAE   & \parbox{2cm}{ \centering Memory usage (MB)}  & \parbox{2cm}{\centering Mean training time}     & \parbox{2cm}{\centering Training time/epoch}  \\ 
\midrule
\parbox{4.5cm}{ARIMA} 
    & 2.1880                       & 1.9005                             & \textemdash         & \textemdash       & \textemdash            \\
\parbox{4.5cm}{ConvLSTM \citep{Shi15}} 
    & 1.8555 $\pm$ 0.0033          & 1.2843 $\pm$ 0.0028                & 922                 & \textbf{02:38:27} & 00:02:21            \\
\parbox{4.5cm}{PredRNN \citep{Wang17}} 
    & 1.6962 $\pm$ 0.0038          & 1.1885 $\pm$ 0.0020                & 2880                & 06:59:34          & 00:05:52            \\
\parbox{4.5cm}{MIM \citep{Wang19}}   
    & 1.6731 $\pm$ 0.0099          & 1.1790 $\pm$ 0.0055                & 4145                & 11:05:37          & 00:10:43            \\
\midrule
STConvS2S-C (ours)        
    & 1.3699 $\pm$ 0.0024          & 0.9434 $\pm$ 0.0020                & 1040                & 03:34:52          & 00:02:48           \\
STConvS2S-R (ours)   
    & \textbf{1.2692} $\pm$ 0.0031 & \textbf{0.8552} $\pm$ 0.0018       & \textbf{895}        & 03:15:12          & \textbf{00:02:13}  \\
\bottomrule
\end{tabular}
\end{adjustbox}

\begin{adjustbox}{width=\textwidth}
\begin{tabular}{lccccc@{\extracolsep{\fill}}} 

                  & \multicolumn{5}{c}{$5 \rightarrow 15$}      \\
\cmidrule{2-6}   
Model       & RMSE    & MAE   & \parbox{2cm}{ \centering Memory usage (MB)}  & \parbox{2cm}{\centering Mean training time}     & \parbox{2cm}{\centering Training time/epoch}  \\ 
\midrule
\parbox{4.5cm}{ARIMA} 
    &  2.2481                       & 1.9077                             & \textemdash         & \textemdash      & \textemdash        \\
\parbox{4.5cm}{ConvLSTM \citep{Shi15}} 
    & 2.0728 $\pm$ 0.0069          & 1.4558 $\pm$ 0.0076                & 1810                & 5:29:30           & 00:07:32            \\
\parbox{4.5cm}{PredRNN \citep{Wang17}} 
    & 2.0237 $\pm$ 0.0067          & 1.4311 $\pm$ 0.0149                & 7415                & 11:45:48          & 00:17:03            \\
\parbox{4.5cm}{MIM \citep{Wang19}}   
    & 2.0287 $\pm$ 0.0361          & 1.4330 $\pm$ 0.0250                & 10673               & 19:19:00          & 00:31:19            \\
\midrule
STConvS2S-C (ours)        
    & 1.8739 $\pm$ 0.0107          & 1.2946 $\pm$ 0.0061                & 1457                & \textbf{03:12:24} & 00:05:17           \\
STConvS2S-R (ours)   
    & \textbf{1.8051} $\pm$ 0.0040 & \textbf{1.2404} $\pm$ 0.0068       & \textbf{1312}       & 03:15:42          & \textbf{00:05:03}  \\
\bottomrule
\end{tabular}

\end{adjustbox}
\end{table}

\subsection{CHIRPS Dataset: results and analysis}
\label{sec:chirps-results}

Similar to what we did with CFSR dataset, we divide the experiments into two phases. The first phase aims to investigate the best hyperparameter settings to adjust the models. In the second, we take into account the best version of each model and perform experiments with different initialization values for weight and bias to consolidate the analysis. In detail, we first set the hyperparameters as in the previous experiments on CFSR dataset. However, as the CHIRPS dataset is almost 4x smaller, all models overfit with those configurations. Thus, as an initial method to address this problem, we reduce the models' complexity by decreasing the number of layers ($L$) and the number of filters ($F$). However, we ensure fair comparability in the way we analyze the learning process when changing $L$ (versions $1$ and $3$), $K$ (versions $2$ and $4$) or $F$ (versions $2$ and $3$). Again, all models were trained mini-batch learning with 50 epochs, and RMSprop optimizer with a learning rate of $10^{-3}$.

\begin{figure}
    \centering
    \begin{minipage}[b]{0.47\linewidth}
        \centering
        \includegraphics[width=\textwidth]{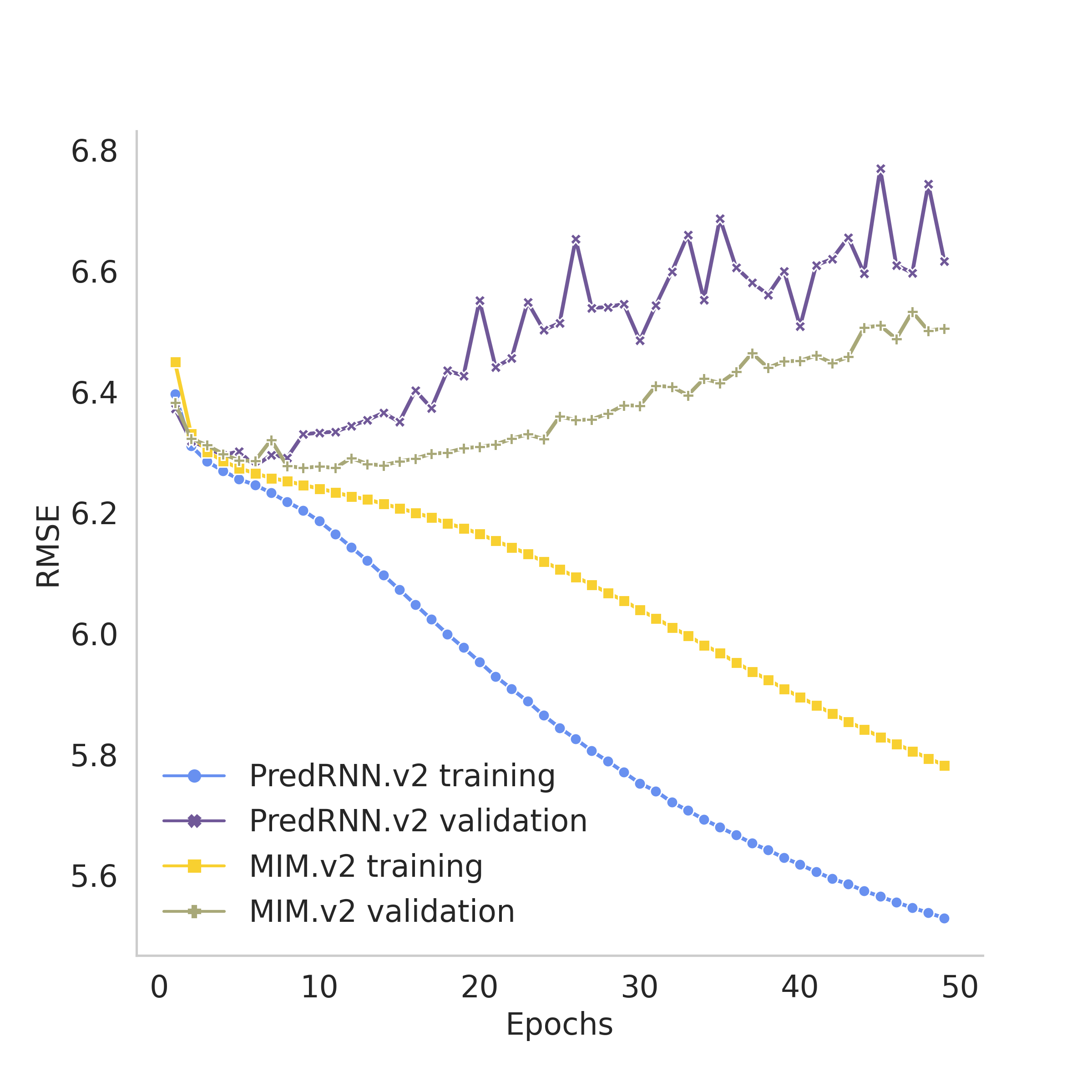}
        \subcaption{Learning curve - overfitting}
    \end{minipage}
    \begin{minipage}[b]{0.47\linewidth}
        \centering
        \includegraphics[width=\textwidth]{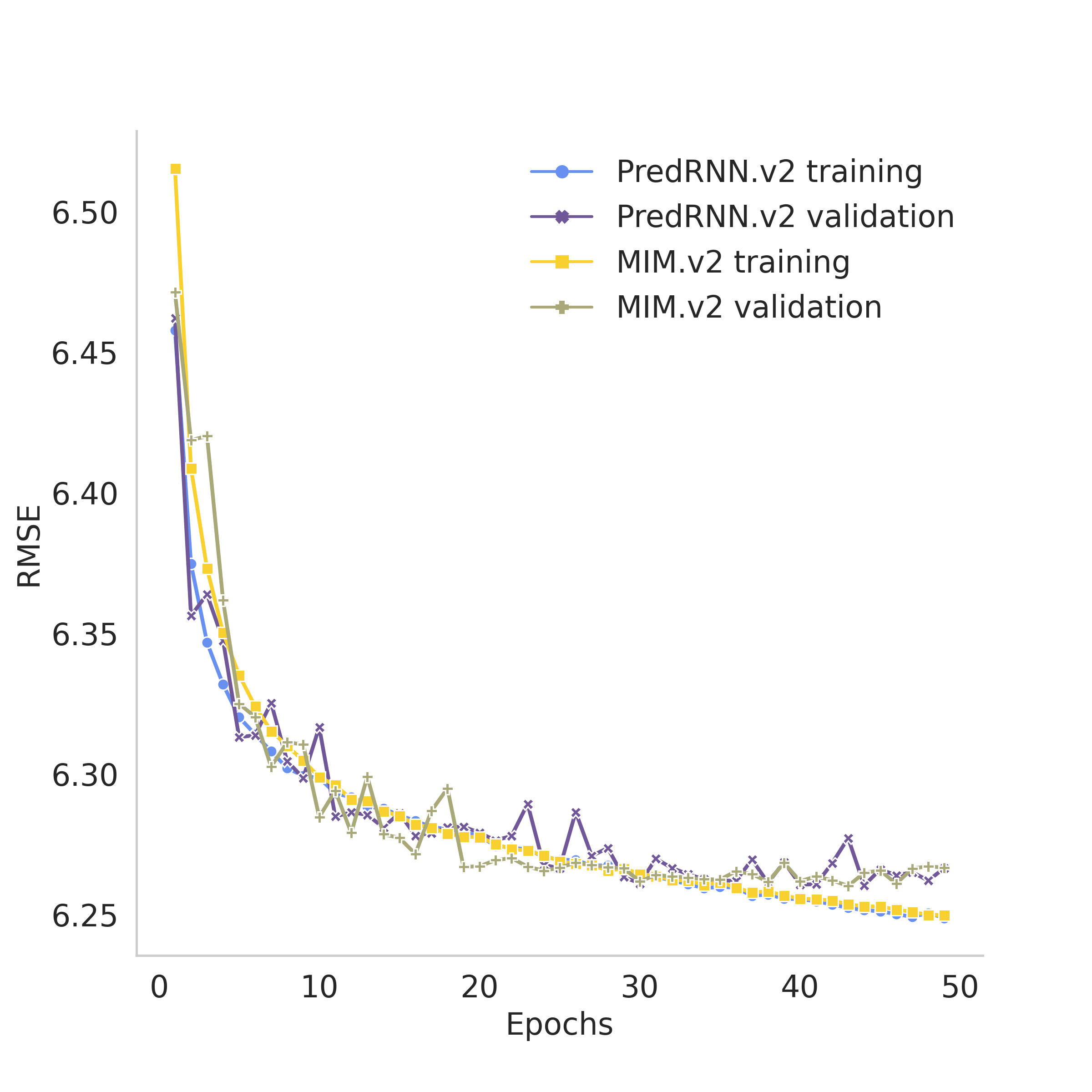}
        \subcaption{Learning curve - dropout 0.5}
    \end{minipage}
        \begin{minipage}[b]{0.5\linewidth}
        \centering
        \includegraphics[width=\textwidth]{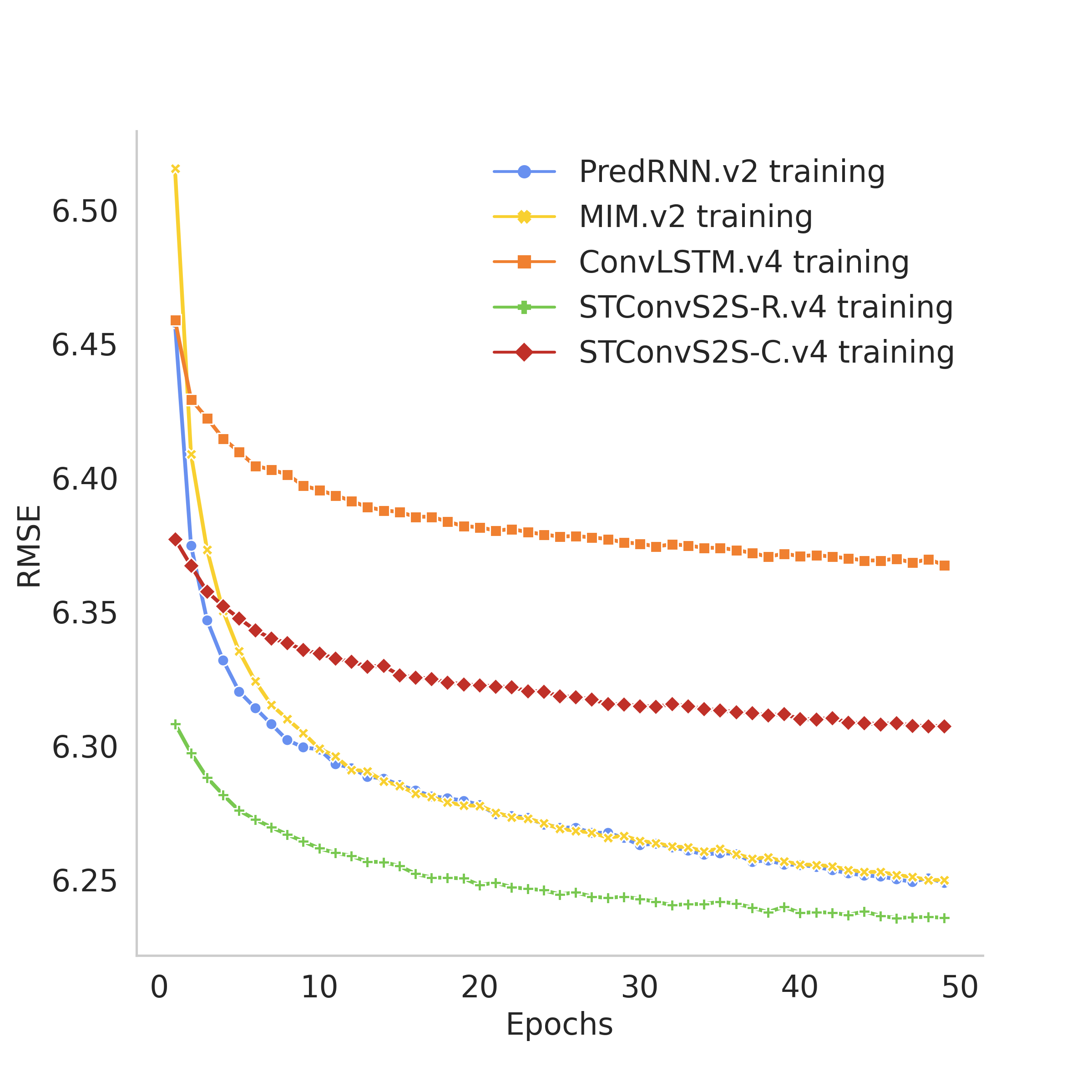}
        \subcaption{Training curve for all models}
    \end{minipage}
    \caption{Learning curves after running 50 epochs on the rainfall dataset (CHIRPS). (a) To exemplify, we select version $2$ to illustrate the overfitting observed when analyzing the PredRNN and MIM models' training and validation curves. (b) The same version and models using dropout to improve its generalization. (c) Comparison of training curve for the best version for each model. STConvS2S-R achieved lower RMSE and, thus, better ability to learn spatiotemporal representations.}
    \label{fig:chirps-lineplot}
\end{figure}

Although we reduced the complexity, the overfitting problem remained with PredRNN and MIM models. We apply dropout to improve their performance in comparison with our proposed models. As before, we apply a search to find the best dropout rate among {0.2, 0.5, 0.8}. Figure \ref{fig:chirps-lineplot} (a) and (b) show the learning curves of these models with overfitting and with a dropout rate of 0.5 applied, respectively. Table \ref{tab:chirps-exp1} shows the experimental results of predicting five grids into the future by observing five grids ($5 \rightarrow 5$). For all models, version $1$ with the fewest layers has the lowest memory usage and was faster in training than the other versions. Another notable analysis is that version $2$ an $4$ consume the GPU memory equally, except for the STConvS2S-C model, thus increasing the kernel size affects only computational time.

\begin{figure}
    \centering
    \includegraphics[width=0.7\linewidth]{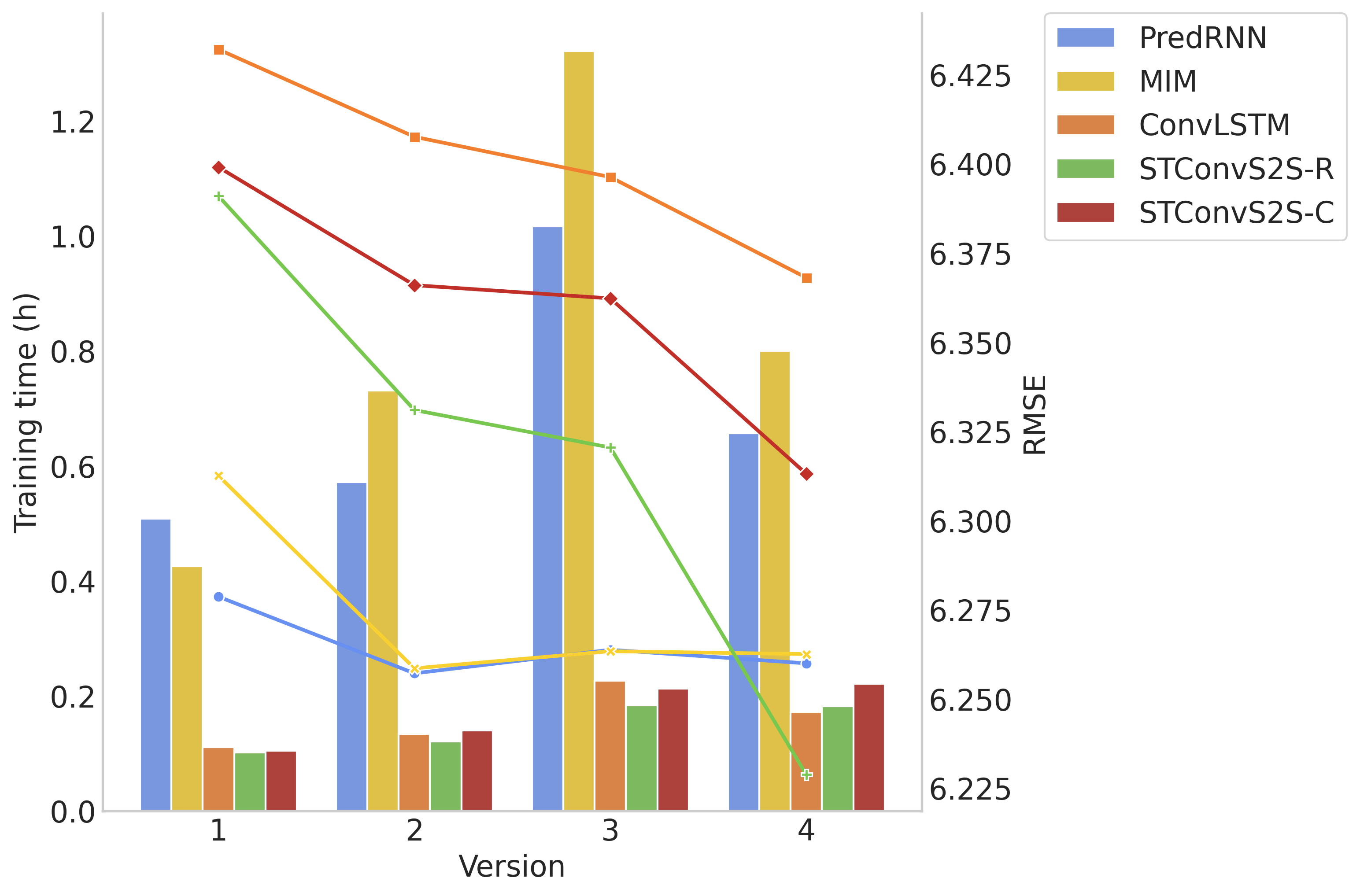}
    \caption{Comparison between training time in hours (bar plot) and RMSE (line plot) for each model version on the rainfall dataset (CHIRPS).}
    \label{fig:chirps-rmse-all-models}
\end{figure}

Figure \ref{fig:chirps-rmse-all-models} compares these results version by version. STConvS2S models outperform ConvLSTM with compatible training time on all versions. Comparing the best version of each model (version $4$ for STConvS2S models and ConvLSTM, and version$2$ for PredRNN and MIM), STConvS2S-R has the lowest prediction error and, compared to PredRNN, it is 3x faster and occupies only 1/3 of memory space. Besides, Figure \ref{fig:chirps-lineplot} (c) illustrates training in 50 epochs and indicates that STConvS2S-R learns the spatiotemporal representation of rainfall better than RNN-based architectures.

For the second phase, we train the models in the same set up as previously indicated for the CFSR dataset. We also include ARIMA methods as a baseline and evaluate the proposed architectures and state-of-the-art models in two tasks: feeding only five observations (grids) into the network and predicting the next 5 and 15 observations, denoted as $5 \rightarrow 5$ and $5 \rightarrow 15$, respectively. For ARIMA, predictions were performed throughout all the 2,500 time series, considering the previous five values in the sequence in each analysis. Results of Table \ref{tab:chirps-exp2} demonstrate that STConvS2S-R achieves a better trade-off between computational cost and prediction accuracy than state-of-the-art models in both tasks. 

\begin{figure}[htb]
    \centering
    \begin{minipage}[b]{1\linewidth}
        \centering
        \includegraphics[width=0.7\textwidth]{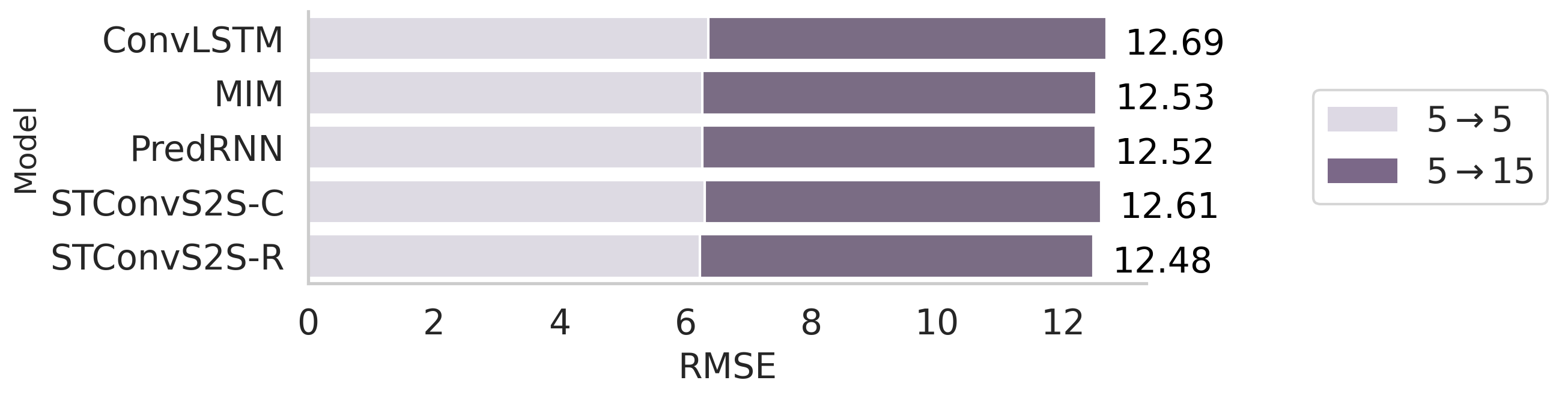}
    \end{minipage}
    \hspace{0.5cm}
    \begin{minipage}[b]{1\linewidth}
        \centering
        \includegraphics[width=0.7\textwidth]{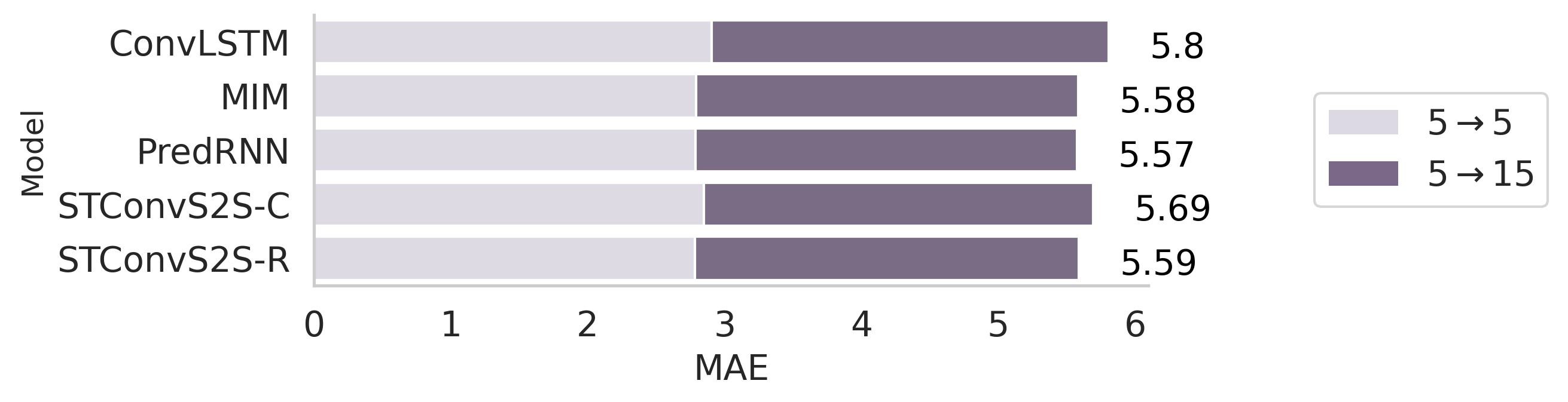}
    \end{minipage}
    \caption{Cumulative error based on both horizons ($5 \rightarrow 5$ and $5 \rightarrow 15$) using the rainfall dataset (CHIRPS). Evaluations on RMSE and MAE metrics.}
    \label{fig:stackedplot-chirps}
\end{figure}

\begin{figure*}
    \includegraphics[width=\linewidth]{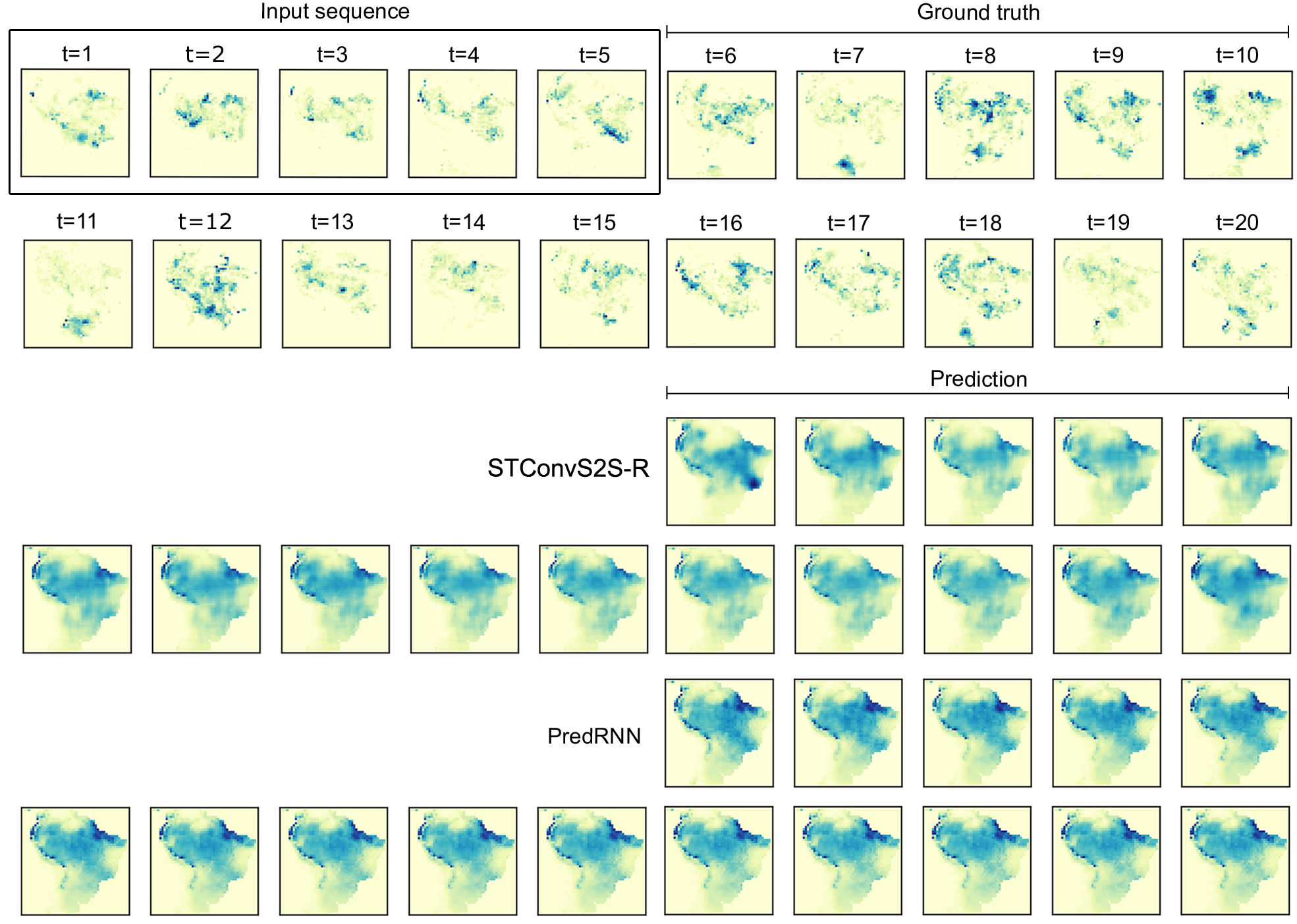}
    \caption{Prediction example on test set of rainfall dataset (CHIRPS). Comparison between the best CNN-based and RNN-based models: STConvS2S-R and PredRNN, respectively}
    \label{fig:grids-chirps}
\end{figure*}

Figure \ref{fig:stackedplot-chirps} summarizes these results in an overview of the cumulative error based on the two forecast horizons. Trained on the rainfall dataset, our proposed architecture equipped with the temporal reversed block achieves performance comparable to RNN-based architecture. Besides, it can predict short and even long sequences in the spatiotemporal context. Such a statement is confirmed by Figure \ref{fig:grids-chirps}, which shows the observations at each time step for STConvS2S-R and PredRNN models. STConvS2S-R can predict in the long-term without many distortions since it presents a predictive result similar to the PredRNN. Given the high variability of rainfall, both models have difficulties in making an accurate forecast.

\begin{table}
\centering
\caption{Evaluation of different settings on the CHIRPS dataset for STConvS2S and state-of-the-art methods, where the best version has the lowest RMSE value.}
\label{tab:chirps-exp1}
\begin{tabular}{lccccc@{\extracolsep{\fill}}} 
\toprule
                                  &         &                  & \multicolumn{3}{c}{$5 \rightarrow 5$}                         \\
\cmidrule{4-6}                            
Model         & Version & Setting   & RMSE              & Training time     &  \parbox{2.5cm}{\centering Memory usage (MB)}      \\ 
\midrule
\multirow{4}{*}{\parbox{3cm}{ConvLSTM \\ \citep{Shi15}}} 
    & 1       & L=1, K=3, F=16    & 6.4321              & \textbf{00:06:39}  & \textbf{746}         \\
    & 2       & L=2, K=3, F=8     & 6.4076              & 00:08:01           & 756                  \\ 
    & 3       & L=2, K=3, F=16    & 6.3963              & 00:13:35           & 989                  \\
    & 4       & L=2, K=5, F=8     & \textbf{6.3681}     & 00:10:19           & 756                  \\

\midrule
\multirow{4}{*}{\parbox{3cm}{PredRNN \\ \citep{Wang17}}}      
    & 1       & L=1, K=3, F=16    & 6.2787              & \textbf{00:30:31}  & \textbf{1673}      \\
    & 2       & L=2, K=3, F=8     & \textbf{6.2572}     & 00:34:19           & 1740                \\ 
    & 3       & L=2, K=3, F=16    & 6.2638              & 01:01:00           & 2775                \\
    & 4       & L=2, K=5, F=8     & 6.2600              & 00:39:24           & 1740                 \\
\midrule
\multirow{4}{*}{\parbox{3cm}{MIM \\ \citep{Wang19}}}      
    & 1       & L=1, K=3, F=16    & 6.3126              & \textbf{00:25:32}  & \textbf{1447}                \\
    & 2       & L=2, K=3, F=8     & \textbf{6.2586}     & 00:43:52           & 2231       \\ 
    & 3       & L=2, K=3, F=16    & 6.2634              & 01:19:18           & 3521                \\
    & 4       & L=2, K=5, F=8     & 6.2626              & 00:48:00           & 2231                \\
\midrule
\multirow{4}{*}{\parbox{3cm}{STConvS2S-C (ours)}}      
    & 1       & L=1, K=3, F=16    & 6.3991              & \textbf{00:06:16}  & \textbf{609}        \\
    & 2       & L=2, K=3, F=8     & 6.3660              & 00:08:23           & 654                \\ 
    & 3       & L=2, K=3, F=16    & 6.3623              & 00:12:45           & 807                \\
    & 4       & L=2, K=5, F=8     & \textbf{6.3131}     & 00:13:16           & 662                 \\
\midrule
\multirow{4}{*}{\parbox{3cm}{STConvS2S-R (ours)}}      
    & 1       & L=1, K=3, F=16    & 6.3910              & \textbf{00:06:05}  & \textbf{584}        \\
    & 2       & L=2, K=3, F=8     & 6.3310              & 00:07:14           & 616                 \\ 
    & 3       & L=2, K=3, F=16    & 6.3205              & 00:11:01           & 735                \\
    & 4       & L=2, K=5, F=8     & \textbf{6.2288}     & 00:10:55           & 616                 \\
\bottomrule
\end{tabular}
\end{table}

\begin{table}
\centering
\caption{Performance results for rainfall forecasting using the previous five observations (grids) to predict the next five observations ($5 \rightarrow 5$), and the next 15 observations ($5 \rightarrow 15$).}
\label{tab:chirps-exp2}
\begin{adjustbox}{width=\textwidth}
\begin{tabular}{lccccc@{\extracolsep{\fill}}} 
\toprule
                  & \multicolumn{5}{c}{$5 \rightarrow 5$}      \\
\cmidrule{2-6}   
Model       & RMSE    & MAE   & \parbox{2cm}{ \centering Memory usage (MB)}  & \parbox{2cm}{\centering Mean training time}     & \parbox{2cm}{\centering Training time/epoch}  \\ 
\midrule
\parbox{4.5cm}{ARIMA} 
    & 7.4377                       & 6.1694                             & \textemdash         & \textemdash       & \textemdash            \\
\parbox{4.5cm}{ConvLSTM \citep{Shi15}} 
    & 6.3666 $\pm$ 0.0019          & 2.9074 $\pm$ 0.0185                & 752                 & \textbf{00:15:15} & 00:00:13            \\
\parbox{4.5cm}{PredRNN \citep{Wang17}} 
    & 6.2625 $\pm$ 0.0039          & 2.7880 $\pm$ 0.0110                & 1740                & 00:39:59          & 00:00:43            \\
\parbox{4.5cm}{MIM \citep{Wang19}}   
    & 6.2621 $\pm$ 0.0051          & 2.7900 $\pm$ 0.0178                & 2231                & 00:52:13          & 00:00:52            \\
\midrule
STConvS2S-C (ours)        
    & 6.3091 $\pm$ 0.0029          & 2.8487 $\pm$ 0.0280                & 662                 & 00:15:54          & 00:00:15           \\
STConvS2S-R (ours)   
    & \textbf{6.2248} $\pm$ 0.0006 & \textbf{2.7821} $\pm$ 0.0261       & \textbf{616}        & 00:16:48          & \textbf{00:00:13}  \\
\bottomrule
\end{tabular}
\end{adjustbox}

\begin{adjustbox}{width=\textwidth}
\begin{tabular}{lccccc@{\extracolsep{\fill}}} 

                  & \multicolumn{5}{c}{$5 \rightarrow 15$}      \\
\cmidrule{2-6}   
Model       & RMSE    & MAE   & \parbox{2cm}{ \centering Memory usage (MB)}  & \parbox{2cm}{\centering Mean training time}     & \parbox{2cm}{\centering Training time/epoch}  \\ 
\midrule
\parbox{4.5cm}{ARIMA} 
    &   7.9460                     & 5.9379                             & \textemdash         & \textemdash      & \textemdash        \\
\parbox{4.5cm}{ConvLSTM \citep{Shi15}} 
    & 6.3244 $\pm$ 0.0025          & 2.8972 $\pm$ 0.0264                & 1308                & 00:44:30          & \textbf{00:00:33}  \\
\parbox{4.5cm}{PredRNN \citep{Wang17}} 
    & 6.2600 $\pm$ 0.0013          & \textbf{2.7850} $\pm$ 0.0067       & 4115                & 01:53:30          & 00:01:58            \\
\parbox{4.5cm}{MIM \citep{Wang19}}   
    & 6.2722 $\pm$ 0.0020          & 2.7935 $\pm$ 0.0246                & 5276                & 02:48:38          & 00:02:34            \\
\midrule
STConvS2S-C (ours)        
    & 6.2962 $\pm$ 0.0039          & 2.8452 $\pm$ 0.0130                & 916                 & \textbf{00:35:43} & 00:00:41           \\
STConvS2S-R (ours)   
    & \textbf{6.2590} $\pm$ 0.0023 & 2.8054 $\pm$ 0.0175                & \textbf{912}        & 00:39:35          & 00:00:39  \\
\bottomrule
\end{tabular}

\end{adjustbox}
\end{table}

\subsection{Ablation Study}
\label{sec:ablation}
We conduct a series of ablation studies, where the goal is to understand our architecture by removing/changing its main components and observing the impact on the evaluation metrics. For fair comparisons, we trained all models with the same settings of version $4$ for CFSR (see Table \ref{tab:cfsr-exp1}) and CHIRPS (see Table \ref{tab:chirps-exp1}) datasets. Besides analyzing the structure of our model, we also compare it with three models: vanilla 3D CNN, 3D Encoder-Decoder architecture \citep{Racah17}, and a CNN model using (2+1)D blocks \citep{Tran18}. In Table \ref{tab:cfsr-ablation} and \ref{tab:chirps-ablation}, the results of these comparisons are shown on rows 1-3, rows 4-11 show the ablation experiments and on rows 12-13 the proposed models in this work. Following, we discuss the experimental results in detail.

\begin{itemize}
\item \textbf{Removal of the factorized convolutions.} In these experiments, there is no separation in temporal and spatial blocks in our architecture, since this split is only possible due to factorized convolutions. These models in relation to layers are similar to 3D CNN (see Figure \ref{fig:architecture-comparison} (a) and (c)). Removing factorized convolutions from the models underperform the proposed models in RMSE and MAE metrics (see rows 4 and 12; rows 5 and 13). To reinforce that the performance gain comes from design options rather than increased model parameters, we also performed experiments removing the progressive filters from our models (rows 10-11) for comparison. The models without factorized convolutions still performed worse in most cases.

\item \textbf{Removal of the causal constraint.} In general, respecting the temporal order is more a restriction of the problem domain than an additional feature to improve models' performance. However, the STConvS2S-R model slightly improves the results, at least in one of the evaluation metrics on both datasets (comparison between rows 6 and 13). This enhancement can also be observed in 3D CNN and "not factorized" STConvS2S-R (rows 1 and 5), as both use the same layers, but differ concerning the causal constraint. On the other hand, the STConvS2S-C model results do not show the same contribution to performance.

\item \textbf{Removal of the temporal block.} This experiment analyzes the importance of this component in our network since we propose two variations of STConvS2S based on the temporal block adopted. The results on row 7 indicate that although faster than the proposed models, this removal has a critical impact on RMSE and MAE, especially for the CFSR dataset.

\item \textbf{Inverted blocks.} Understanding the influence of the temporal and spatial blocks on each other is not straightforward. Thus, to analyze the model structure, we change the blocks from (temporal $\Rightarrow$ spatial) to (spatial $\Rightarrow$ temporal). Both GPU memory usage and training time are very similar, comparing each STConvS2S model with the respective inverted version. Regarding the evaluation metrics, the inverted versions have the worst performance on both datasets, except for MAE on CHIRPS using STConvS2S-R. 

\item \textbf{Comparison with baselines methods.} There is no significant differentiation among STConvS2S models and the baselines on the CHIRPS dataset concerning memory usage and training time. These metrics almost increase twice on the CFSR dataset compared to 3D CNN and 3D Encoder-Decoder, but as a result, STConvS2S-R achieves a 4\% improvement in RMSE over those same baselines. STConvS2S-R had a favorable or matching performance in the results of the experiments on both datasets, except for the MAE metric on the CHIRPS dataset comparing against (2+1)D Conv. STConvS2S-C does not perform as well as STConvS2S-R in these comparisons.

\item \textbf{Comparison of strategies to satisfy the causal constraint.} The results show the superiority of STConvS2S-R compared to STConvS2S-C in the evaluation metrics and time performance in all the experiments. The hypothesis of STConvS2S-R to better predict is that this model adds less zero-padding before performing the convolution operation on each layer inside the temporal block. Concerning time efficiency, in STConvS2S-R, the reverse function is performed only twice in the temporal reversed block, making it faster than STConvS2S-C. The latter performs its operations on each layer within the temporal causal block (Section~\ref{subsec:temporal-block}). This study demonstrates the relevance of our original method to make convolutional layers satisfy the causal constraint during the learning process.
\end{itemize}

\begin{table}
\centering
\caption{Quantitative comparison of ablation experiments, baseline methods using 3D convolutional layers, and our proposed models on temperature dataset  (CFSR) for $5 \rightarrow 5$ task. }
\label{tab:cfsr-ablation}
\begin{adjustbox}{width=\textwidth}
\begin{threeparttable}
\begin{tabular}{lcccccc@{\extracolsep{\fill}}} 
\toprule
\qquad Model & \parbox{2cm}{\centering Factorized Conv.}  & \parbox{1.5cm}{ \centering Causal const.} & RMSE  & MAE & \parbox{2.5cm}{ \centering Memory usage (MB)}  & \parbox{2cm}{\centering Training time}  \\
\midrule
\parbox{4cm}{\rownumber.  \space  3D CNN}                 
                                            & \textemdash     & \textemdash          & 1.3307     & 0.9015      &   578  & 01:13:54 \\
\parbox{4cm}{\rownumber. \space  3D Encoder-Decoder}      
                                            & \textemdash     & \textemdash          & 1.3327     & 0.9291      &   544  & 01:11:26 \\
\parbox{4cm}{\rownumber. \space  (2+1)D Conv}
                                            & \checkmark      & \textemdash          & 1.2944     & 0.8763      &  847   & 01:51:24 \\
\midrule
\parbox{4cm}{\rownumber. \space STConvS2S-C}     
                                            & \textemdash     & \checkmark           & 1.4450     & 1.0068      &   605  & 01:39:47 \\
\parbox{4cm}{\rownumber. \space STConvS2S-R}  
                                            & \textemdash     & \checkmark           & 1.3215     & 0.8958      &   580  & 01:16:53 \\
\parbox{4cm}{\rownumber. \space STConvS2S}       
                                            & \checkmark      & \textemdash          & 1.2811     & 0.8645      &   884  & 02:00:52 \\
\parbox{4cm}{\rownumber. \space STConvS2S\tnote{*}}  
                                            & \checkmark      & \textemdash          & 1.6780     & 1.1828      &   740  & 01:14:47 \\
\parbox{4cm}{\rownumber. \space STConvS2S-C\tnote{**}} 
                                             & \checkmark     & \checkmark          & 1.4152     & 0.9750      &   1000  & 02:15:46 \\
\parbox{4cm}{\rownumber. \space STConvS2S-R\tnote{**}}
                                             & \checkmark     & \checkmark          & 1.3044     & 0.8796      &   895  & 01:56:05 \\
\parbox{4cm}{\rownumber.  STConvS2S-C\tnote{***}} 
                                             & \checkmark     & \checkmark          & 1.4218     & 0.9821      &   698  & 01:04:58 \\
\parbox{4cm}{\rownumber.  STConvS2S-R\tnote{***}} 
                                             & \checkmark     & \checkmark          & 1.3234     & 0.8966      &   649  & 00:57:11 \\
\midrule
\parbox{4cm}{\rownumber.  STConvS2S-C} & \checkmark     & \checkmark          & 1.3791     & 0.9492      &   1040  & 02:25:26    \\
\parbox{4cm}{\rownumber.  STConvS2S-R} & \checkmark     & \checkmark          & 1.2773     & 0.8646      &   895   & 01:58:39    \\
\bottomrule
\end{tabular}
\begin{tablenotes}
    \small
    \item[*] No temporal block 
    \item[**] Inverted (spatial $\Rightarrow$ temporal)
    \item[***] No filter increase 
\end{tablenotes}
\end{threeparttable}
\end{adjustbox}
\end{table}

\setcounter{magicrownumbers}{0}

\begin{table}
\centering
\caption{Quantitative comparison of ablation experiments, baseline methods using 3D convolutional layers, and our proposed models on rainfall dataset  (CHIRPS) for $5 \rightarrow 5$ task. }
\label{tab:chirps-ablation}
\begin{adjustbox}{width=\textwidth}
\begin{threeparttable}
\begin{tabular}{lcccccc@{\extracolsep{\fill}}} 
\toprule
\qquad Model & \parbox{2cm}{\centering Factorized Conv.}  & \parbox{1.5cm}{ \centering Causal const.} & RMSE  & MAE & \parbox{2.5cm}{ \centering Memory usage (MB)}  & \parbox{2cm}{\centering Training time}  \\
\midrule
\parbox{4cm}{\rownumber.  \space  3D CNN}                 
                                            & \textemdash     & \textemdash          & 6.2519     & 2.8519      &   534  & 00:09:27 \\
\parbox{4cm}{\rownumber. \space  3D Encoder-Decoder}      
                                            & \textemdash     & \textemdash          & 6.2540     & 2.7977      &   513  & 00:09:26 \\
\parbox{4cm}{\rownumber. \space  (2+1)D Conv}
                                            & \checkmark      & \textemdash          & 6.2323     & 2.7243      &   660  & 00:12:09 \\
\midrule
\parbox{4cm}{\rownumber. \space STConvS2S-C}     
                                            & \textemdash     & \checkmark           & 6.3310     & 2.9161      &   553  & 00:12:38 \\
\parbox{4cm}{\rownumber. \space STConvS2S-R}  
                                            & \textemdash     & \checkmark           & 6.2510     & 2.8082      &   534  & 00:09:55 \\
\parbox{4cm}{\rownumber. \space STConvS2S}       
                                            & \checkmark      & \textemdash          & 6.2281     & 2.8134      &   609  & 00:10:19 \\
\parbox{4cm}{\rownumber. \space STConvS2S\tnote{*}}  
                                            & \checkmark      & \textemdash          & 6.3539     & 2.8980      &   572  & 00:06:57 \\
\parbox{4cm}{\rownumber. \space STConvS2S-C\tnote{**}} 
                                             & \checkmark     & \checkmark          & 6.3255     & 2.8594      &   656  & 00:12:10 \\
\parbox{4cm}{\rownumber. \space STConvS2S-R\tnote{**}}
                                             & \checkmark     & \checkmark          & 6.2397     & 2.7971      &   616  & 00:10:56 \\
\parbox{4cm}{\rownumber.  STConvS2S-C\tnote{***}} 
                                             & \checkmark     & \checkmark          & 6.3171     & 2.8418      &   591  & 00:10:23 \\
\parbox{4cm}{\rownumber.  STConvS2S-R\tnote{***}} 
                                             & \checkmark     & \checkmark          & 6.2434     & 2.7829      &   565  & 00:09:25 \\
\midrule
\parbox{4cm}{\rownumber.  STConvS2S-C} & \checkmark     & \checkmark          & 6.3131     & 2.8327      &   662  & 00:13:16 \\
\parbox{4cm}{\rownumber.  STConvS2S-R} & \checkmark     & \checkmark          & 6.2288     & 2.8060      &   616  & 00:10:55 \\
\bottomrule
\end{tabular}
\begin{tablenotes}
    \small
    \item[*] No temporal block 
    \item[**] Inverted (spatial $\Rightarrow$ temporal)
    \item[***] No filter increase 
\end{tablenotes}
\end{threeparttable}
\end{adjustbox}
\end{table}

\section{Conclusion} \label{sec:conclusion}
Predicting future information many steps ahead can be challenging, and a suitable sequence-to-sequence architecture to better represent spatiotemporal data for this purpose is still open for research. However, RNN-based models have been widely adopted in these cases \citep{Shi15,Wang17, Wang19}. Previously to this work, CNN-based architectures were not considered for this task, due to two limitations. Firstly, they do not respect the temporal order in the learning process. They also cannot generate an output sequence of length that is higher than the input sequence. Considering these limitations, we proposed STConvS2S, an end-to-end trainable deep learning architecture. STConvS2S can do spatiotemporal data forecasting by only using 3D convolutional layers.

First, we address the problem of causal constraint, proposing two variations of our architecture, one using the temporal causal block and the other, the temporal reversed block. The former adopts causal convolution, which is commonly used in 1D CNN. We also introduced a new technique in the temporal reversed block that makes it not violate the temporal order by applying a reverse function in the sequence. These implementations are essential for a fair comparison with state-of-the-art methods, which are causal models due to the chain-like structure of LSTM layers. To overcome the sequence length limitation, we designed a temporal generator block at the end of the architecture to extend the spatiotemporal data only in the time dimension. 
We further compared our models with state-of-the-art models through experimental studies in terms of both performance and time efficiency on meteorological datasets. The results indicate that our model manages to analyze spatial and temporal data dependencies better since it has achieved superior performance in temperature forecasting, and comparable results in the rainfall forecasting, with the advantage of being up to 5x faster than RNN-based models. Thus, STConvS2S could be a natural choice for sequence-to-sequence tasks when using spatiotemporal data. We evaluate our architecture in the weather forecasting problem, but it is not limited to this domain. We expect that the results presented in this work will foment more research with comparisons between convolutional and recurrent architectures. 

For future work, we will search for ways to decrease rainfall dataset error. Directions may include applying preprocessing techniques to sparse data and adding data from other geographic regions. Besides, we will investigate more architectures for spatiotemporal data forecasting.

\section*{Computer Code Availability}
We implemented the deep learning models presented in this paper using PyTorch 1.0, an open-source framework. Our source code is publicly available at \url{https://github.com/MLRG-CEFET-RJ/stconvs2s}

\section*{Data Availability}
In this paper, spatiotemporal datasets in NetCDF format were used and can be downloaded at \url{http://doi.org/10.5281/zenodo.3558773}, an open-source online data repository.

\section*{Acknowledgment}

The authors thank CNPq, CAPES, FAPERJ, and CEFET/RJ for partially funding this research.

\bibliographystyle{elsarticle-harv}
\bibliography{main}  

\end{document}